\documentclass[10pt,journal,compsoc]{IEEEtran}

\ifCLASSOPTIONcompsoc
  \usepackage[nocompress]{cite}
\else
  \usepackage{cite}
\fi

\usepackage[pdftex]{graphicx}
\usepackage{amsmath}
\usepackage{bbm}
\usepackage[capitalize]{cleveref}
\usepackage{url}
\usepackage[shortlabels]{enumitem}
\usepackage{wrapfig}
\usepackage{subcaption}
\usepackage{float}

\usepackage{algorithm}
\usepackage{algpseudocode}
\algrenewcommand\algorithmicrequire{\textbf{input}}

\newcommand{\indep}{\perp \!\!\! \perp}
\newcommand{\veryshortarrow}[1][3pt]{\mathrel{%
   \vcenter{\hbox{\rule[-.5\fontdimen8\textfont3]{#1}{\fontdimen8\textfont3}}}%
   \mkern-4mu\hbox{\usefont{U}{lasy}{m}{n}\symbol{41}}}}

\usepackage{amsmath,amsfonts,bm}

\def\eqref#1{equation~\ref{#1}}

\def\1{\bm{1}}

\def\va{{\bm{a}}}

\DeclareMathAlphabet{\mathsfit}{\encodingdefault}{\sfdefault}{m}{sl}
\SetMathAlphabet{\mathsfit}{bold}{\encodingdefault}{\sfdefault}{bx}{n}

\newcommand{\E}{\mathbb{E}}

\begin{document}  
\bstctlcite{IEEEexample:BSTcontrol}

\IEEEtitleabstractindextext{

\begin{abstract}
The Gumbel-max trick is a method to draw a sample from a categorical distribution, given by its unnormalized (log-)probabilities. Over the past years, the machine learning community has proposed several extensions of this trick to facilitate, e.g., drawing multiple samples, sampling from structured domains, or gradient estimation for error backpropagation in neural network optimization. The goal of this survey article is to present background about the Gumbel-max trick, and to provide a structured overview of its extensions to ease algorithm selection. Moreover, it presents a comprehensive outline of (machine learning) literature in which Gumbel-based algorithms have been leveraged, reviews commonly-made design choices, and sketches a future perspective.
\end{abstract}

}

\title{A Review of the Gumbel-max Trick and its Extensions for Discrete Stochasticity\\ in Machine Learning}
\author{
\thanks{
© 2022 IEEE.  Personal use of this material is permitted.  Permission from IEEE must be obtained for all other uses, in any current or future media, including reprinting/republishing this material for advertising or promotional purposes, creating new collective works, for resale or redistribution to servers or lists, or reuse of any copyrighted component of this work in other works.}
Iris A. M. Huijben,~\IEEEmembership{Student Member,~IEEE}, Wouter Kool,\\ Max B. Paulus, Ruud J. G. van Sloun,~\IEEEmembership{Member,~IEEE}
\IEEEcompsocitemizethanks{\IEEEcompsocthanksitem I.A.M Huijben is affiliated with the department
of Electrical Engineering, Eindhoven University of Technology, The Netherlands, and Onera Health, Eindhoven, The Netherlands. E-mail: i.a.m.huijben@tue.nl
\IEEEcompsocthanksitem R.J.G. van Sloun is affiliated with the department of Electrical Engineering, Eindhoven University of Technology, The Netherlands, and Philips Research, Eindhoven, The Netherlands. Email: r.j.g.v.sloun@tue.nl
\IEEEcompsocthanksitem W. Kool is affilitated with the Amsterdam Machine Learning Lab (AMLab), University of Amsterdam, The Netherlands, and ORTEC R\&D, Zoetermeer, The Netherlands. E-mail: w.w.m.kool@uva.nl
\IEEEcompsocthanksitem M.B. Paulus is affiliated with the department of Computer Science, ETH, Zurich, Switzerland. E-mail: max.paulus@inf.ethz.ch}
}


\maketitle

\section{Introduction}

The world around us is discrete in many aspects. Think about decision making, e.g. in traffic (\textit{Should I decelerate, accelerate or maintain a constant speed?}), for product selection (\textit{Given that I liked the trousers from shop X last time, which new trousers shall I buy?}), or in a clinical setting (\textit{Should I administer medication A or B to the patient?}). Other discrete examples yield occurrence of events (\textit{At which day did I see you for the last time?}), social networks (\textit{Do two persons know each other?}) or data compression (\textit{How many bits do we need to store this information?}). Modelling these concepts evidently pleads for discrete models, on which we focus in this work.

The immensely grown popularity of machine learning, and in particular deep learning, has given rise to high capacity models that have consistently been outperforming more classical models in various domains. Whereas the first neural networks typically comprised a stack of few fully-connected layers, activated with non-linear functions, current deep learning models often exhibit a (very) deep modular architecture. Specifically generative models (e.g. the variational auto-encoder (VAE) \cite{kingma2013} and auto-regressive pixelCNN \cite{oord2016}), and Bayesian models combine such deep architectures with a form of stochasticity. Trainable parameters of these models are generally learned in a data-driven fashion, i.e. they are optimized over a set of training examples by back-propagating the error for a downstream task. The probability distributions are then treated as stochastic nodes in a computation graph, from which a sample is drawn to compute the output, and subsequently the error.

Sampling from resulting discrete distributions may raise challenges when the probability mass function is unnormalized, or in case of an (exponentially) large sampling domain. The latter is often encountered in sequence models, where one `sample' entails a full sequence, and the sampling domain grows combinatorially for each additional sequence element. Moreover, incorporating (discrete) stochasticity in deep learning models poses a second challenge. Gradient computation, and therefore error backpropagation, through a stochastic node is hampered, which is required for updating the distribution's parameters and all model parameters that precede this node in the network. These two challenges, i.e. \emph{sampling} and \emph{gradient estimation} of discrete stochasticity in deep learning models, will be the focus of this review.

Several algorithms today exist to sample from structured models with exponentially large sampling domains (e.g. ancestral sampling), or to sample from a set of unnormalized probabilities (e.g. Markov Chain Monte Carlo (MCMC) methods \cite{hastings1970monte}, or the Gumbel-max trick \cite{Gumbel1954}). The Gumbel-max trick recently found renewed attention for use in deep learning models, thanks to the proposed Gumbel-Softmax (GS) gradient estimator that is based on a relaxation of this trick \cite{Jang2017,Maddison2017}. The GS estimator (and variants thereof) have become popular (biased) alternatives for the high-variance REINFORCE estimator \cite{Williams1992}, thanks to their good empirical performance and straightforward implementation.

This article provides both an intuitive and mathematical understanding of the Gumbel-max trick, reviews extensions of this trick, provides handles for algorithm selection and corresponding design choices, and sketches a future perspective. The content can be summarized as follows:
\begin{itemize}
    \item Background on categorical random variables, the Gumbel distribution and inverse transform sampling is covered in \cref{sec:preliminaries}.
    \item Applications of Gumbel-based algorithms in machine learning are discussed in \cref{sec:applications}.
    \item Section \ref{sec:sampling_algorithms} and \ref{sec:gradient_estimation_NN} present technical details about Gumbel-based sampling algorithms and gradient estimators, respectively. 
    \item Practical considerations and commonly-made design choices are presented in \cref{sec:considerations}.
    \item Section \ref{sec:conclusion} summarizes this review and sketches a future perspective.
\end{itemize}

\section{Preliminaries}
\label{sec:preliminaries}

\subsection{Categorical distribution}
\label{sec:cat_dist}
A categorical distribution is a probability distribution that assigns a probability to $N$ distinct classes. We consider three different parameterizations of the same distribution: \emph{normalized probabilities} $\boldsymbol{\pi}$, \emph{unnormalized probabilities} $\boldsymbol{\theta}$, or \emph{unnormalized log-probabilities} (or \emph{logits}) $\log \boldsymbol{\theta} = \va/T$, where $T \in \mathbb{R}_{>0}$ is a temperature parameter. We denote the $i^{\text{th}}$ unnormalized probability - which depends on $T$ - with $\theta_{i;T}$. The probabilities $\pi_{i;T}$ for each class $i \in D=\{1,\ldots,N\}$ are then given by:
\begin{equation}
    \label{eq:normprobs_vs_unnorm_logits}
    \pi_{i;T} = \frac{\theta_{i;T}}{\sum_{j \in D} \theta_{j;T}} = \frac{\exp(a_i/T)}{\sum_{j \in D} \exp(a_j/T)} = \frac{\exp(a_i/T)}{Z_D(\boldsymbol{\theta})},
\end{equation}
where $Z_D(\boldsymbol{\theta}) = \sum_{j\in D}\theta_{j;T} = \sum_{j \in D}\exp(a_j/T) \in \mathbb{R}_{>0}$ is the \emph{partition function} that normalizes the distribution. To prevent clutter, both the parameter and subscript of $Z$, and temperature subscript $T$ are, in the rest of this paper, omitted when context allows it.

The temperature $T$ controls the distribution's entropy, such that the distribution can vary between a degenerate/deterministic `one-hot' distribution for $T\rightarrow0^{+}$ (i.e. all mass is centralized in one class), and a Uniform distribution (all probabilities are equal) for $T \rightarrow \infty$. In many cases, $T$ simply equals one, making $\va$ the unnormalized log-probabilities. Equation (\ref{eq:normprobs_vs_unnorm_logits}) is known as a \emph{softmax} function with temperature parameter $T$. Parameterized as $\operatorname{Cat}(\va,T)$, the categorical distribution is often referred to as the Gibbs or Boltzmann distribution.

We will use the terms \emph{unstructured} and \emph{structured} models in this work. The former is used for what is normally referred to as a categorical distribution: a distribution assigning a probability to a single event/class (e.g. the probability that horse $X$ wins a horse race). The number of distribution parameters thus equals the size of the sampling domain. A structured model, on the other hand, places a probability on structures, i.e. different combinations of events/classes, rather than on single events (e.g. the probability that horses $X$, $Y$, and $Z$ cover the top-3). A sequence model is another common example of such a structured model, in which the sampling distribution of each element may be conditional upon the previously sampled element.

A random variable following a categorical distribution is denoted with $I$ in this work, and a realization (i.e. a sample from the categorical distribution) is denoted with $I = \omega$, providing the \emph{index} of the sampled class. In certain contexts, it is useful to define this sample as a one-hot embedding; a unit vector of length $N$, with a one at index $\omega$ and zeros otherwise, which we denote with $\mathbbm{1}_{\omega}$. Several algorithms exist to sample from a categorical distribution. Inverse transform sampling is the most basic approach (see \cref{sec:standard_sampling}). The Gumbel-max trick (see \cref{sec:gumbel_max}), and variants thereof (see \cref{sec:extended_sampling_algorithms}) are commonly-used alternatives in machine learning applications.

\subsection{Gumbel distribution}
\label{sec:gumbel_dist}

The Gumbel distribution \cite{Gumbel1935} is an instance (type I) of the generalized extreme value distribution\footnote{Also the Fréchet (type II) and Weibull (type III) distribution generalize to this generalized extreme value distribution.} \cite{mises1936}, which models optima and rare events. A Gumbel random variable - which is often referred to in this work as `a Gumbel' - is parameterized by location and scale parameters $\mu \in \mathbb{R}$ and $\beta \in \mathbb{R}_{\geq 0}$, respectively. The corresponding probability (PDF) and cumulative density (CDF) functions are respectively given by
\begin{align}
    \label{eq:gumbel_pdf}
    f(x) &= \frac{1}{\beta}e^{-\frac{x-\mu}{\beta}} e^{-e^{-\frac{x-\mu}{\beta}}} , \hspace{1cm} \mathrm{and}\\
    F(x) &= e^{-e^{-\frac{x-\mu}{\beta}}}.
\end{align}

We denote a Gumbel distribution (as defined in \cref{eq:gumbel_pdf}) with $\operatorname{Gumbel}(\mu,\beta)$ (or sometimes $\operatorname{G}(\mu,\beta)$ in short), and a random variable following this distribution with $G_{\mu,\beta}$. 
To prevent clutter, the scale, or both parameters are frequently omitted when standard settings are assumed (i.e. $G_{\mu} := G_{\mu,1}$ and $G := G_{0,1}$). We often consider a set of identically and independently distributed (i.i.d.) Gumbel variables, where we use $G^{(i)}$ to denote an $i^{\text{th}}$ standard Gumbel. Note that, $\mu$ and $\beta$ are \emph{not} the mean and variance of a Gumbel, which are given by:
\begin{align}
\label{eq:gumbel_mean}
&\mathbb{E}[\operatorname{G}_{\mu,\beta}] = \mu + \gamma\beta, \hspace{2cm} \mathrm{and}\\
&\mathbb{E}[(\operatorname{G}_{\mu,\beta}-\mathbb{E}[\operatorname{G}_{\mu,\beta}])^2] = \frac{\pi^2}{6}\beta^2,
\end{align}
where $\gamma \approx 0.577$ is the Euler constant and $\pi \approx 3.14$ is the constant pi (not to be confused with the normalized probability of the categorical distribution), respectively.

The inverse cumulative density function (ICDF; also called quantile function) is given by
\begin{equation}
\label{eq:gumbel_icdf}
    F^{-1}(u) = -\beta\log(-\log u)+\mu.
\end{equation}
From \cref{eq:gumbel_icdf} it can be seen that the Gumbel distribution is closed under \emph{scaling and addition}, i.e. any Gumbel variable can be generated by scaling and shifting a standard Gumbel. 

Equation (\ref{eq:gumbel_icdf}) is used in inverse transform sampling (see \cref{sec:standard_sampling}) to transform a sample from the Uniform distribution $\operatorname{U}(0,1)$ into a Gumbel sample via a double (negative) logarithmic relation. Taking only one negative logarithm of a Uniform random variable, i.e. $X = -\log U$, defines a sample from the (standard) Exponential distribution. The Gumbel distribution is, therefore, sometimes also referred to as the `Double Exponential' distribution. Thanks to the aforementioned relations, some properties of the Gumbel distribution are closely related to properties of the Exponential, and Uniform/Beta distribution, on which we elaborate in Section \ref{sec:uni_exp_gumbel_section}.

\begin{figure}
    \centering
    \includegraphics[width=\linewidth,trim={0cm 16cm 16.3cm 1.5cm},clip]{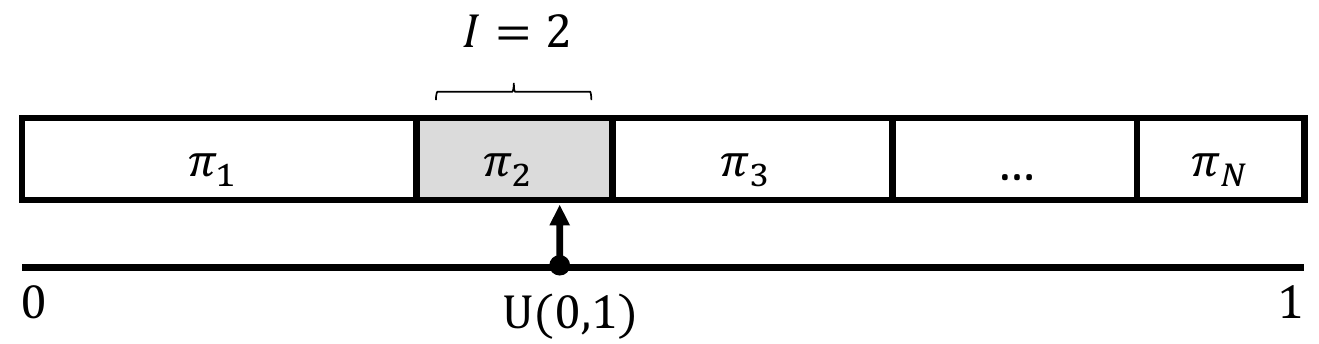}
    \caption{Illustration of inverse transform sampling for drawing a sample from $\operatorname{Cat}(\boldsymbol{\pi})$. A sample from the Uniform distribution is converted to one of the categories/classes of the categorical distribution via its inverse cumulative density function. The chance that class $i$ is being sampled equals $\pi_i$; the width of the corresponding box in this illustration.}
    \label{fig:inverse_transform}
\end{figure}

\subsection{Inverse transform sampling}
\label{sec:standard_sampling}
Inverse transform sampling is the most common way to sample from a distribution. It transforms a standard Uniform variable into another random variable, via a transformation characterized by the random variable's ICDF. Figure \ref{fig:inverse_transform} illustrates this process for sampling from a categorical distribution: a sample form a Uniform random variable is inserted into the ICDF of the categorical distribution in order to find the corresponding bin/class. Inverse transform sampling is easy to implement and only requires generation of one random variate per categorical sample. It does, however, require the categorical distribution to be normalized. The ICDF of the Gumbel distribution (given in \cref{eq:gumbel_icdf}) can equivalently be used to draw a sample from this distribution.

\section{Applications}
\label{sec:applications}

In machine learning, there is significant interest in over-parameterized modular and structured models, which often involve one or more stochastic components. When concerned with \emph{discrete} stochasticity, challenges arise regarding both \emph{sampling} from discrete distributions, and \emph{gradient estimation} thereof.

Sampling from discrete distributions can be achieved with (among others) inverse transform sampling, or the Gumbel-max trick \cite{Gumbel1954} (see \cref{sec:gumbel_max}) and extensions thereof (see \cref{sec:extended_sampling_algorithms}). Gumbel-based sampling algorithms have, for example, been used for (discrete) action selection in a multi-armed bandit setting \cite{Cesa-Bianchi2017}, for sampling data points in active learning \cite{Ma2020}, for text generation in dialog systems \cite{Firdaus2020}, or in translation tasks \cite{Kool2019a,Kool2020a}.

\begin{figure}
\centering
\includegraphics[width=\linewidth,trim={0cm 1.4cm 12.5cm 1.5cm},clip]{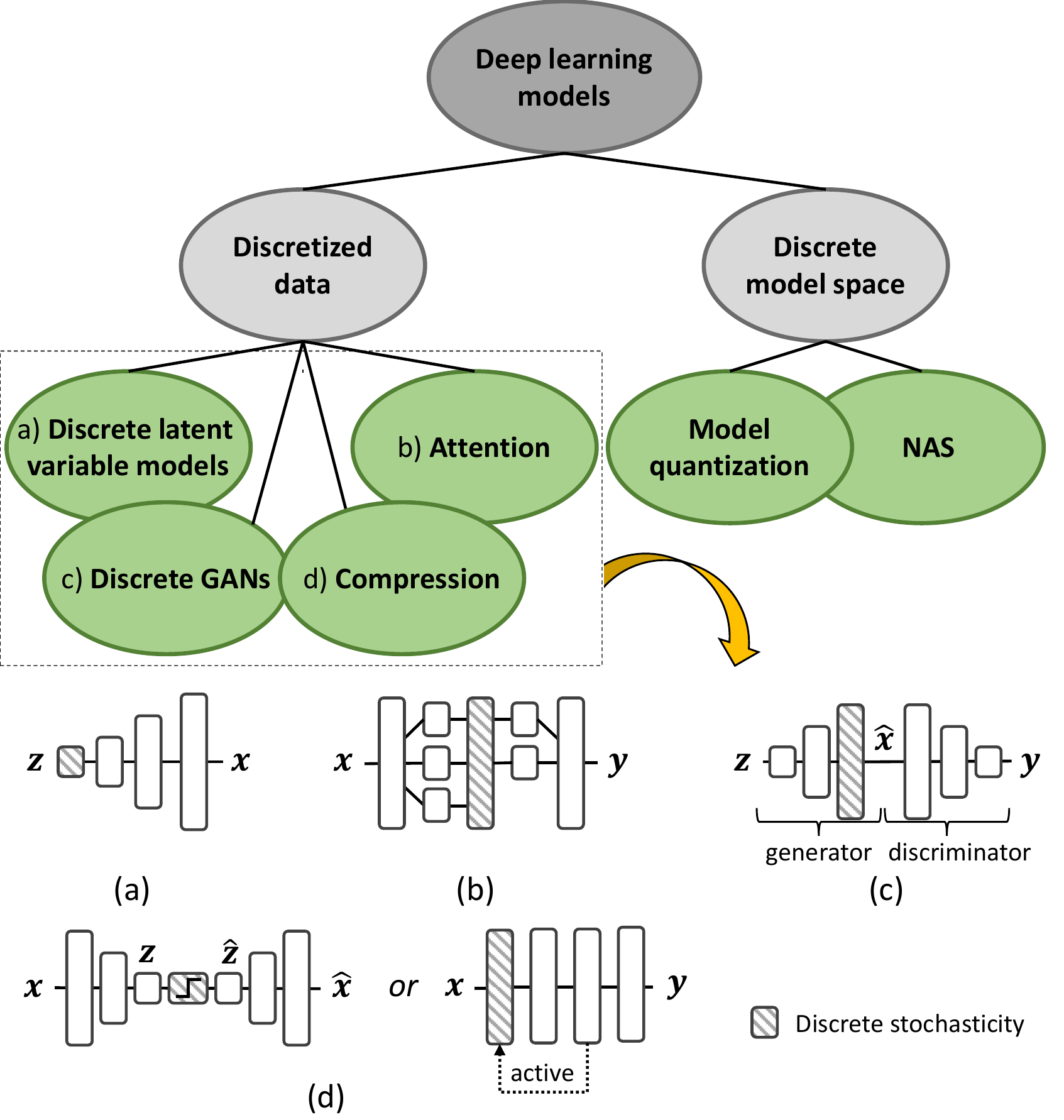}
\caption{Categorization of the different applications in deep learning in which Gumbel-based gradient estimators have been applied in order to facilitate training via backpropagation through discrete stochastic nodes. Visualizations in the bottom show where different models exhibit discretized data, indicated by the dashed gray boxes.}
\label{fig:application_tree}
\end{figure}

When the stochastic components are parameterized, and the model is trained end-to-end with stochastic gradient-based methods, gradients need to be taken through the stochastic sampling process and gradient estimators are therefore called for. Finding such estimators is particularly challenging when the stochastic components are discrete. Relaxations of the (non-differentiable) Gumbel-max trick, notably the Gumbel-Softmax (GS) estimator and variants therefore, have been found useful for this purpose, and will be discussed in \cref{sec:gradient_estimation_NN}. In this section we review the different applications, and refer to any variant of the GS estimator as a Gumbel-based estimator, in order to provide a general overview without elaborating on technical details. Applications are categorized into works that leverage discretized data, or perform model selection from a discrete (model) space (see \cref{fig:application_tree}).

\subsection{Discretized data}
We interpret the term \emph{discretized data} in a wide sense here. This category includes models that rely on discrete representations (e.g. categorical probability distributions in discrete latent variable models), attention mechanisms (sub-selecting data is a discrete choice process), generative models for inherently discrete data (e.g. GANs for text generation), and models for data compression.

\subsubsection{Discrete latent variable models}

Latent variable models assume that `the world' (represented as training data) has originated from a set of latent (i.e. hidden) variables. While such variables are often modelled continuously, many situations arise in which the choice for discrete latent variables seems more appropriate (see \cref{fig:application_tree}a).
Such discrete latent variable models have mainly come in two flavours. On the one hand, one can have a discrete approximation/representation of a continuous (latent) variable (i.e. quantization, as in \cite{van2017neural}), while on the other hand, one could model the data with a discrete (non-degenerate) distribution in the latent space. In the former case the straight-through estimator \cite{bengio2013STE} is typically used to facilitate gradient updates of the encoder. The latter case, on the other hand, requires sampling from the introduced discrete distribution to estimate the expectation in the forward pass and its gradient during backpropagation. This approach is also possible in quantization settings, as will be discussed in \cref{sec:application_data_compression}. All following works in the current sub-section have adopted the second approach and leveraged Gumbel-based estimators for discrete latent variable models.

VAEs with discrete priors are proposed by \cite{Silveira2018,Lievin2019,ramesh2021zero}, and the authors of
\cite{Liu2018a} relax the discrete latents and durations of a recurrent hidden semi-Markov Model. In some works, continuous and discrete latent variables have been combined \cite{Figueroa2017, Dupont2018, Figueroa2019, Wang2019c}. The authors of \cite{Figueroa2017} leverage - next to a continuous latent variable - (only) one categorical random variable, which is then interpreted as a clustering variable that assigns data points to discrete classes, resulting in a combined classification- and generative model. Similarly, \cite{Figueroa2019} train a VAE with a Gaussian Mixture model as prior, where assignment to either of the Gaussians displays a discrete process as well, and in the same line, the authors of \cite{Chen2018,Gao2020} relax backpropagation of their deep clustering algorithms.  Applications of (Gumbel-based) discrete latent variable models as described above include (among others) planning \cite{Asai2017}, syntactic parsing \cite{Corro2018}, text modelling \cite{Baynazarov2019a}, speech modelling \cite{Tjandra2019,Baevski2020}, paraphrase generation \cite{Fu2020a}, recommender systems \cite{Li2019a}, drug-drug interaction modelling \cite{Dai2020}, and event modelling \cite{Rezaee2021}.

\subsubsection{Attention}
\label{sec:attention}
Attention mechanisms are concerned with filtering incoming information, analogously to the way we - human beings -  selectively observe the world around us. Hard attention resorts to either accepting or declining information, while soft attention refers to placing more emphasis on certain parts than on others. When training neural networks, only soft attention allows gradients to flow, so hard (i.e. discrete) attention mechanisms have been relaxed using Gumbel-based gradient estimators to enable model optimization. While the term attention is typically used to denote selectivity in hidden representations of deep learning models, we slightly widen the term and present works that select part of the data (either hard or soft) anywhere in the model (see \cref{fig:application_tree}b), with the main motivation to improve the model's performance or its interpretability. Attention has been used on latent features to enhance interpretability of `black box' neural networks \cite{Chen2018a,Abid2019}, for hierarchical multi-scale recurrent neural networks \cite{Yan2018}, and in graph neural networks (GNNs) \cite{Kang2019,Acharya2020}. The authors of \cite{Acharya2020} apply attention on edges with the specific goal of graph clustering. Applications range from (but are not limited to) recommender systems \cite{Tay2018}, pose estimation \cite{Zhou2019}, video classification \cite{Huang2019}, event detection \cite{Ngo2020}, and image synthesis \cite{Tseng2020}. The authors of \cite{Yang2020} apply attention on features of discrete input samples and an attacking vocabulary, in order to acquire a scalable method for real-time generation of adversarial examples on discrete input data. Hard attention on (input) data points (i.e. learned subsampling) has, moreover, been leveraged to reduce computational overhead of neural networks, e.g. on point clouds \cite{Yang2019a} and graphs \cite{BhaskarAcharya2020}. 

Also decision making in an environment of one or multiple agents can be considered (hard) attention. The authors of \cite{Niu2019} leverage this `attention' to enable end-to-end training of an agent that makes discrete decisions. In a multi-agent environment, such decisions have concerned communication symbol selection \cite{Havrylov,Mordatch2017} in order to learn a language to be used among agents.

\subsubsection{Discrete GANs}
Generative Adversarial Networks (GANs) \cite{goodfellow2014generative} have been a popular class of generative models. They contain a generator and discriminator module (see \cref{fig:application_tree}c). The latter's aim is to distinguish between real and fake (generated by the generator) samples. This setup requires differentiability of the discriminator loss with respect to the generator's parameters. However, in the case of applying a GAN on discrete data, the discrete and stochastic fashion of the generator's output hampers direct differentiability, and has therefore been a suitable candidate for Gumbel-based gradient estimators. Applications of discrete GANs that have leveraged these Gumbel-based gradient estimators include (among others) text generation \cite{Kusner2016, Xu2017, Shetty, Lu2017, Gu2018, Nie2019, Kojima2021}, fake user data creation for recommender systems \cite{Chonwiharnphan2020}, and action prediction \cite{Zhao2020}. Discrete GANs have also been combined with knowledge distillation frameworks. The authors of \cite{Baram2017}, for example, propose to learn a policy (i.e. a trajectory of subsequent states and actions in reinforcement learning setups) by learning to imitate a (known) expert/teacher policy. The student model learns this imitation by fooling a discriminator that should distinguish between both policies. Such setups may, however, require a large number of training iterations before the student model converges. To alleviate this, the authors of \cite{Wang2019a} have proposed the concept of adversarial distillation in the context of multi-label classification, in which both the teacher and student model are jointly battling against a discriminator that should distinguish their (discrete) label predictions from real annotations. Both models learn from each other via distillation losses, finally resulting in a low-resource student model that can be used during inference. Both the teacher and student model's output are relaxed using the GS estimator.

\subsubsection{Data compression}
\label{sec:application_data_compression}

Our growing demand for data has sky-rocketed data rates, hampering data transfer, processing and storage. To alleviate these rates, data compression comes into play. A shift is currently being made from non-data-driven compression techniques (e.g. JPEG) to data-driven methods, where parameters of a neural compression model are learned from a training set of data. Such models typically encode the data in a quantized and lower-dimensional space, after which the data are decoded again (see \cref{fig:application_tree}d). Quantization is a discrete process, making it a suitable candidate to be learned using a Gumbel-based estimator. The authors of \cite{Shu2018, Wei2020} propose to learn the quantization levels, which form a quantization codebook, whereas the authors of \cite{Kim2020} learn to select such a codebook as a whole. Moreover, data-adaptive binarization has been learned as well \cite{Tsai2018,Zhang,Yang2020a}. 

The amount of acquired data in the digital domain is governed by the analog-to-digital (ADC) converter in the sensing stage. Instead of compressing in the digital domain, the compressed sensing (CS) framework \cite{eldar2012compressed} introduces compression directly at this sensing stage (see \cref{fig:application_tree}d). Simply reducing the sampling frequency of an ADC introduces aliasing artifacts in case the minimal Nyquist rate is not met. CS leverages incoherent sampling patterns (expressed in a sensing matrix) to beat this famous Nyquist rate under certain assumptions. The typically heuristic (and task-independent) design of this sensing matrix, has been outperformed by a task-conditional (discrete) sensing matrix, learned via a Gumbel-based estimator \cite{Huijben2020a}. Once trained, such learned sensing system can be implemented in hardware to reduce data rates at the sensing stage directly. The proposed framework has been applied for medical ultrasound imaging \cite{Huijben2019}, magnetic resonance imaging \cite{Huijben2020}, and MIMO antenna arrays \cite{Yang2020c}.  Moreover, it was extended to a data-conditional setup, by making the sensing matrix conditional upon already acquired data \cite{vanGorp2021}. Note that data compression by subsampling can also be considered (hard) attention, as discussed in \cref{sec:attention}. Although the focus of this data compression section is on works that aim for data reduction rather than model enhancement. 

\subsection{Discrete model space}
The number of available neural models has given rise to algorithms that not only optimize a model on training data, but also search for a suitable model itself. Such data-driven model selection can be leveraged with the aim to just find better-fitting models, or to find compressed models that facilitate implementation in dedicated hardware. We distinguish methods that learn concepts related to model quantization, and methods that learn hyper-parameters (often related to depth and width). The latter is often referred to as neural architecture search (NAS) or pruning. 

\subsubsection{Model quantization}
Model quantization can take place on weights, activations and/or gradients. Joint optimization and quantization requires differentiability of the quantization operation, making it a suitable candidate for Gumbel-based gradient estimators. Indeed, it was shown that quantization of network weights and activations can effectively be learned, by either learning the quantization grid while using a pre-defined number of bits \cite{Louizos2018a}, or learning the appropriate number of bits per layer \cite{Wu2018}. The authors of \cite{Lacey2018} also learn layer-dependent bits assignment by leveraging a bits budget to be distributed, for quantization of weights, activations and gradients.

\subsubsection{Neural architecture search (NAS) \& pruning}
Gumbel-based algorithms for discrete NAS of deep neural networks have been leveraged for distinct goals. Most general, hyper-parameters have been learned - rather than set heuristically - to alleviate the burden of tuning \cite{Gal2017,Chang2019,Xie2018}. In case such hyper-parameters relate to the size of the model (e.g. width or depth), NAS (often called pruning in this case) is generally used with the aim to find small models that facilitate implementation in (dedicated) hardware \cite{Louizos2018b,Grathwohl2018a,Wu2019,Dong2019,Herrmann2020,Wan2020,Kang2020,gonzalez2021}. Moreover, models have been proposed in which the functionality is conditioned upon incoming data. This conditioning can be applied to improve performance \cite{Li2018,Herrmann2020,Jen-TzungChien2019}, achieve run-time acceleration (i.e. reduce computational complexity during inference) \cite{Veit2017,Bejnordi2019,Wang2020a,Verelst2020,Cai2021}, or enhance interpretability \cite{Li2018, Liu2018b,Schlichtkrull2021}. Also multi-task models have been subject to discrete NAS, where task-dependent gating/branching has been learned \cite{Bragman2019,Guo2020}. Considering graphs as structured data models, NAS has also been leveraged to find suitable graphical representations, e.g., by optimizing the adjacency matrix \cite{Zhang2019b,Schlichtkrull2021}. Moreover, graph structures are leveraged by Tree-LSTMs \cite{Tai2015}, which were later extended to Gumbel-Tree LSTMs, where graph edges are co-optimized instead of pre-defined \cite{Choi2018a}. Gumbel-Tree LSTMs have notably found use in machine translation \cite{Currey2018, Su2020}.

\section{Sampling algorithms}
\label{sec:sampling_algorithms}

In this section we introduce the Gumbel-max trick and its properties (\cref{sec:gumbel_max}), and link it to concepts known from different fields, i.e. Poisson processes (\cref{sec:uni_exp_gumbel_section}) and Boltzmann exploration (\cref{sec:boltzmann}). We then introduce top-down sampling using the inverted Gumbel-max trick (\cref{sec:gumbel_max_revert}), followed by sampling algorithms that are extensions of the conventional Gumbel-max trick (\cref{sec:extended_sampling_algorithms}).

\subsection{Gumbel-max trick}
\label{sec:gumbel_max_trick}

\subsubsection{Definition and properties}
\label{sec:gumbel_max}

The Gumbel-max trick \cite{Gumbel1954} draws a sample from a categorical random variable $I \sim \operatorname{Cat(\boldsymbol{\pi})}$. It does so, by adding i.i.d. Gumbel (noise) samples to the unnormalized log-probabilities and selecting the index with the maximum value, which in turn follows a Gumbel distribution. More specifically:
\begin{alignat}{4}
    \label{eq:gumbel_max}
&I &{}={}& \underset{i \in D}{\operatorname{argmax}} \{\log \theta_i + G^{(i)} \} &{}\sim{} & \operatorname{Cat}(\boldsymbol{\pi}) \hspace{1.1cm} \mathrm{and} \\
    \label{eq:max_stability}
     &M &{}={}& \underset{i\in D}{\operatorname{max}}\{{\log \theta_i + G^{(i)}}\}  &{}\sim{} & \operatorname{Gumbel}(\log Z).
\end{alignat}

\noindent The arguments $G_{\log \theta_i} := \log \theta_i+G^{(i)}$ ($\forall i \in D$) are shifted independent Gumbels, and often referred to as \emph{perturbed logits}. Equation (\ref{eq:max_stability}) is known as the \emph{max-stability} property, which states that the maximum is \emph{independent} of the argmax random variable: $M \indep I$. The interested reader is referred to Appendix \ref{app:gumbel_max_proof} for the proofs of \cref{eq:gumbel_max} and (\ref{eq:max_stability}). Section \ref{sec:uni_exp_gumbel_section} provides a more intuitive understanding of the Gumbel-max trick, by linking it to properties of well-known Poisson processes.

Scaling (or normalizing) unnormalized probabilities $\boldsymbol{\theta}$ by any positive constant $Z$, results in a subtraction in logarithmic space: $\log (\boldsymbol{\theta} / Z) = \log \boldsymbol{\theta} - \log Z$. 
Given the translation-invariance of the argmax function, we can see that the Gumbel-max trick as defined in \cref{eq:gumbel_max}, also applies for normalized probabilities $\boldsymbol{\pi}$:

$$I = \underset{i \in D}{\operatorname{argmax}}\{\log \theta_i +G^{(i)}\} =
\underset{i \in D}{\operatorname{argmax}}\{\underbrace{~\log \pi_i~}_{\log \theta_i - \log Z} +~G^{(i)}\}.$$

\noindent This property is a direct result of the fact that Gumbel random variables adhere to Luce's choice axiom \cite{Yellott1977}, which states that the probability ratio of selecting two elements from a set is independent of the other elements present in that set. Thanks to this independence, a constant scaling of the probabilities does thus not influence the argmax output. As a result, the Gumbel-max trick can also be used over sub-domain $B \subseteq D$, with unnormalized probabilities $\theta_i$, for $i \in B$ \cite{Maddison2014a}:
$$\underset{i \in B}{\operatorname{argmax}}\{\log \theta_i+G^{(i)}\} \sim \operatorname{Cat}\Big(\frac{\mathbbm{1}(i \in B)~\boldsymbol{\pi}}{\sum_{i \in B}\pi_i}\Big).$$

\noindent In machine learning applications, the parameters of the categorical distribution are often learned or predicted. The Gumbel-max trick is therefore typically used with unnormalized - rather than normalized - probabilities, as this allows for unconstrained optimization of $\log\boldsymbol{\theta}$ (whereas $\boldsymbol{\pi}$ is a probability vector). 

The Gumbel-max trick leverages standard Gumbel noise samples. Nevertheless, when using i.i.d. samples from $\operatorname{Gumbel}(\mu,\beta)$, with $\mu \neq 0$ and $\beta \neq 1 ~(\geq 0)$, the distributions of the sampled index and maximum can still be defined analytically:

\begin{alignat}{4}
    \label{eq:scaled_gumbel_max}
&I' = \underset{i \in D}{\operatorname{argmax}} \{\log \theta_{i;T} + G^{(i)}_{\mu,\beta} \} &{}\sim{} & \operatorname{Cat}\Big(\frac{\exp(\va/(T\beta))}{\underset{i \in D}\sum \exp(a_i/(T\beta))}\Big), \\
    \label{eq:scaled_gumbel_max2}
     &M' = \underset{i\in D}{\operatorname{max}}\{{\log \theta_{i;T} + G^{(i)}_{\mu,\beta}}\}  &{}\sim{} & \operatorname{Gumbel}(\mu+\beta \log Z',\beta),
\end{alignat}

\noindent with $G^{(i)}_{\mu,\beta}$ the $i^{\text{th}}$ independent Gumbel random variable with location $\mu$, and scale $\beta$, and $Z' = \sum_{i \in D} \exp(a_i/(T\beta))$. Appendix \ref{app:gumbel_scaled_max_proof} derives the aforementioned relations. Interestingly, from \cref{eq:scaled_gumbel_max} we conclude that $I' \sim \operatorname{Cat}(\va, T\beta)$. Changing the scale of the independent Gumbel variates thus changes the temperature of the Boltzmann distribution from which exact samples are being drawn (see \cref{sec:boltzmann} for a discussion about this relation). 

\begin{figure*}
    \centering
    \includegraphics[width=\linewidth,trim={0cm 9cm 0cm 2cm},clip]{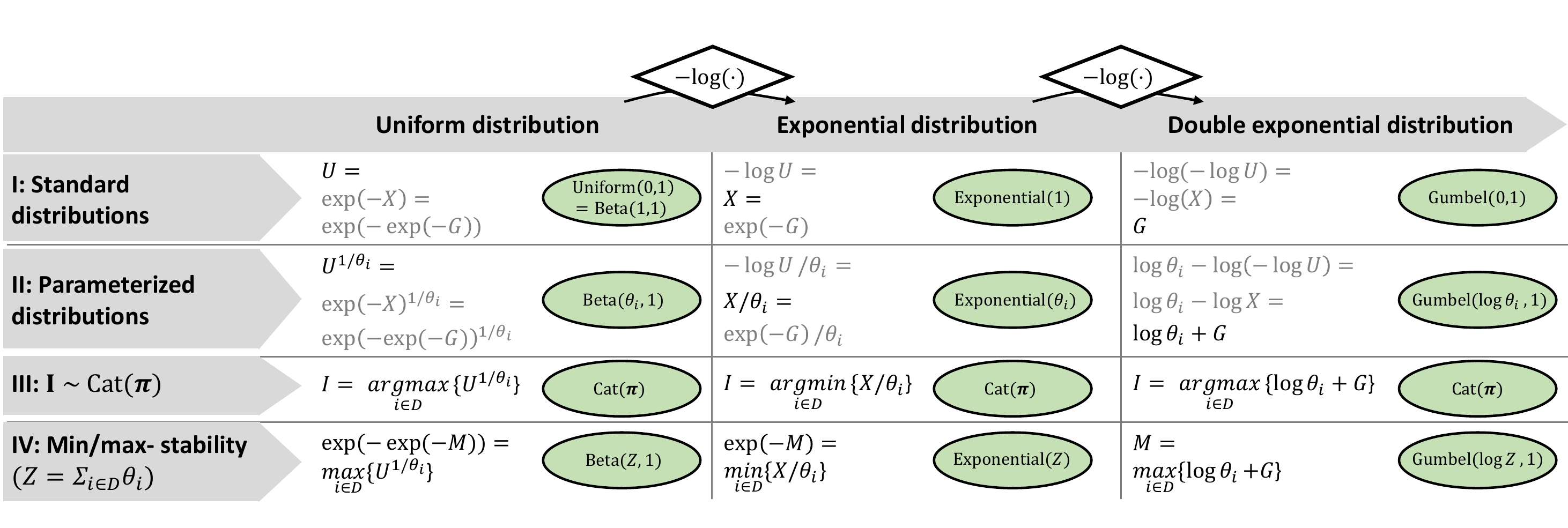}
    \vspace{-0.5cm}
    \caption{The Gumbel (double Exponential) distribution directly relates to the Uniform/Beta and Exponential distribution via a single, respectively double, negative logarithmic relation. The green ellipses indicate the probability distribution of the corresponding random variable in the cell. To prevent clutter, we here define $U := U^{(i)}$, $X := X^{(i)}$, and $G := G^{(i)}$, being an $i^{\text{th}}$ independent standard Uniform, Exponential, and Gumbel random variable, respectively.}
    \label{tab:uni_exp_gumbel}
\end{figure*}

\subsubsection{Link to reservoir sampling and Poisson processes}

\label{sec:uni_exp_gumbel_section}
As mentioned in \cref{sec:gumbel_dist}, the Uniform/Beta, Exponential and Gumbel distribution are closely related, and can be used to express different parameterizations of the same stochastic process. These distinct parameterizations are typically used in different research areas: the Uniform/Beta parameterization is notably used for weighted reservoir sampling \cite{Efraimidis2006}, whereas the Exponential counterpart is widely used in queueing theory/Poisson processes \cite{kingman1992poisson}. The `double Exponential' parameterization is leveraged for the Gumbel-max trick \cite{Gumbel1954}, and its extensions. The aim of this section is to provide a summarizing overview of the existing relations between these distributions, and illustrate this through an intuitive example. Figure \ref{tab:uni_exp_gumbel} relates samples and properties of the three parameterizations (one per column), and will further be explained in this section. Some of the connections that are presented have previously been made, e.g., both the author(s) of \cite{vieira2014gumbel} and \cite{Xie2019} showed the equivalence between weighted reservoir sampling and the Gumbel-max trick, and the author of \cite{Maddison2018} extensively discussed the relation between Poisson processes, especially Exponential races, and the Gumbel-max trick.

The ICDF of the Gumbel distribution - as provided in \cref{eq:gumbel_icdf} - already showed the relation between a Gumbel and a Uniform random variable. The Gumbel distribution also relates to the $\operatorname{Exponential}(\theta)$ distribution ($\theta \in \mathbb{R}_{\geq 0}$), which is characterized by a CDF being $F(x) = 1 - e^{-\theta x}$ for $x \geq 0$ (and 0 for $x < 0$), and location $\mathbb{E}[X] = 1/\theta$. Its ICDF equals $F^{-1}(u) = \frac{-\log(1-u)}{\theta}$. Re-introducing $X \sim \operatorname{Exponential}(1)$ as a standard Exponential random variable with $\theta=1$, allows writing:

\begin{equation}
\label{eq:exp_sample}
\frac{-\log(1-V)}{\theta} \overset{d}{=} \frac{-\log(U)}{\theta} = \frac{X}{\theta},
\end{equation}

\noindent where both $V$ and $U$ are standard Uniform random variables, and $\overset{d}{=}$ denotes equality in distribution. Taking the negative logarithm of \cref{eq:exp_sample}, result in:
\begin{equation}
    \label{eq:transform_exp_sample}
     -\log \frac{X}{\theta} = -\log X + \log \theta = -\log(-\log U) + \log \theta,
\end{equation}

\noindent from which we can see that the negative logarithm transforms an Exponential random variable into a Gumbel random variable, located at $\log \theta$ (right-hand side of \cref{eq:transform_exp_sample}). Note that for $\theta=1$, this Gumbel turns into a standard Gumbel variable with $\mu = 0$. Rows I and II in \cref{tab:uni_exp_gumbel} relate both the standard and parameterized Uniform/Beta, Exponential and Gumbel distributions. The green ellipses indicate the corresponding distributions. 

Besides relating the distributions, we can also link some properties of the Gumbel-max trick to properties of a Poisson process; a stochastic process widely used for modelling the occurrence of events over time. Thanks to the wide study of this process, many readers might have an intuitive understanding about its properties, enabling introduction of these relations via an intuitive example.

The Exponential distribution models the time between sequential events occurring according to a Poisson process. The distribution's parameter $\theta$ represents the intensity (or rate), expressed in events per time unit. As an example, consider a scientific poster presentation where on average 8 students and 2 professors arrive per hour, both according to independent Poisson processes. The Exponential distribution models the time between subsequent arrivals of students and professors (with intensities 8 and 2, respectively). We can sample the arrival process by independently sampling these two Exponentials. From this example we intuitively realize that $P[I=\mathrm{student}] = \frac{\theta_{\mathrm{student}}}{\theta_{\mathrm{student}}+\theta_{\mathrm{prof.}}}$, i.e. the probability that the first person to arrive is a student\footnote{This can be shown by computing $P(X_s < X_p)$ where $X_s$, $X_p$ are the Exponential inter-arrival times for students/professors, respectively.} equals $\frac{8}{8+2} = 0.8$. This property can be linked to the Gumbel-max trick. Formally, given $X^{(i)} \sim \operatorname{Exponential}(1)$ i.i.d. $\forall i \in D$, \cref{eq:transform_exp_sample}, and the monotonicity of the logarithmic function, the following equalities hold:

\begin{align}
\label{eq:gumbel_max_with_exp}
    I &= 
    \underset{i \in D}{\operatorname{argmax}} \{\log \theta_i + G^{(i)} \}
    =\underset{i \in D}{\operatorname{argmax}}\{\log\theta_i-\log X^{(i)}\}  \nonumber \\
    &=\underset{i \in D}{\operatorname{argmin}}\{\underbrace{~X^{(i)}/\theta_i~}_{\sim \operatorname{Exp.}(\theta_i)}\} \sim \operatorname{Cat}(\boldsymbol{\pi}).
\end{align}

\noindent Thus $P[I=\omega] = \pi_{\omega} =  \theta_{\omega} / \sum_{i \in D} \theta_i$, which indeed coincides with our intuitively derived probability \mbox{$P[I = \mathrm{student}]$}. Row III in \cref{tab:uni_exp_gumbel} displays this relation, as well as the relation to the Beta/Uniform distribution. The latter is known from the field of weighted reservoir sampling \cite{Efraimidis2006}, and can directly be deduced from the Exponential reparameterization by taking the negative exponent of the Exponential sample (which, in turn, switches the argmin function to argmax thanks to $1/\exp(\cdot)$ being monotonically decreasing).

Instead of independently sampling the two Exponentials (i.e. one for students and one for professors), we can develop an alternative but equivalent sampling process thanks to the memoryless property of the Exponential distribution\footnote{$P(X > t_1+t_2 | X > t_1) = P(X > t_2)$}. More specifically, we can sample the arrivals of \emph{people} (students and professors) at the poster from an Exponential with an intensity of $8+2=10$ persons per hour, and then independently assign each arrival to either being a student or professor with probabilities 0.8 and 0.2, respectively. The fact that this `merged' Poisson process can be sampled using an Exponential with rate $Z=\sum_{i \in D}\theta_i$, relates to the max-stability property of independent Gumbel variables. Row IV of \cref{tab:uni_exp_gumbel} shows the relation between the three different parameterizations. The actual value of the optimum in the Exponential parameterization equals the shortest arrival time, i.e. the time at which the first person arrives. Note that this time does \emph{not} provide any information about the person being a student or a professor, i.e. the index and value of the optimum are independent, analogously to the independence between $I$ and $M$ in the Gumbel-max trick.

\begin{figure*}
    \centering
    \includegraphics[width=1\linewidth,trim={0cm 2.3cm 0cm 1.5cm},clip]{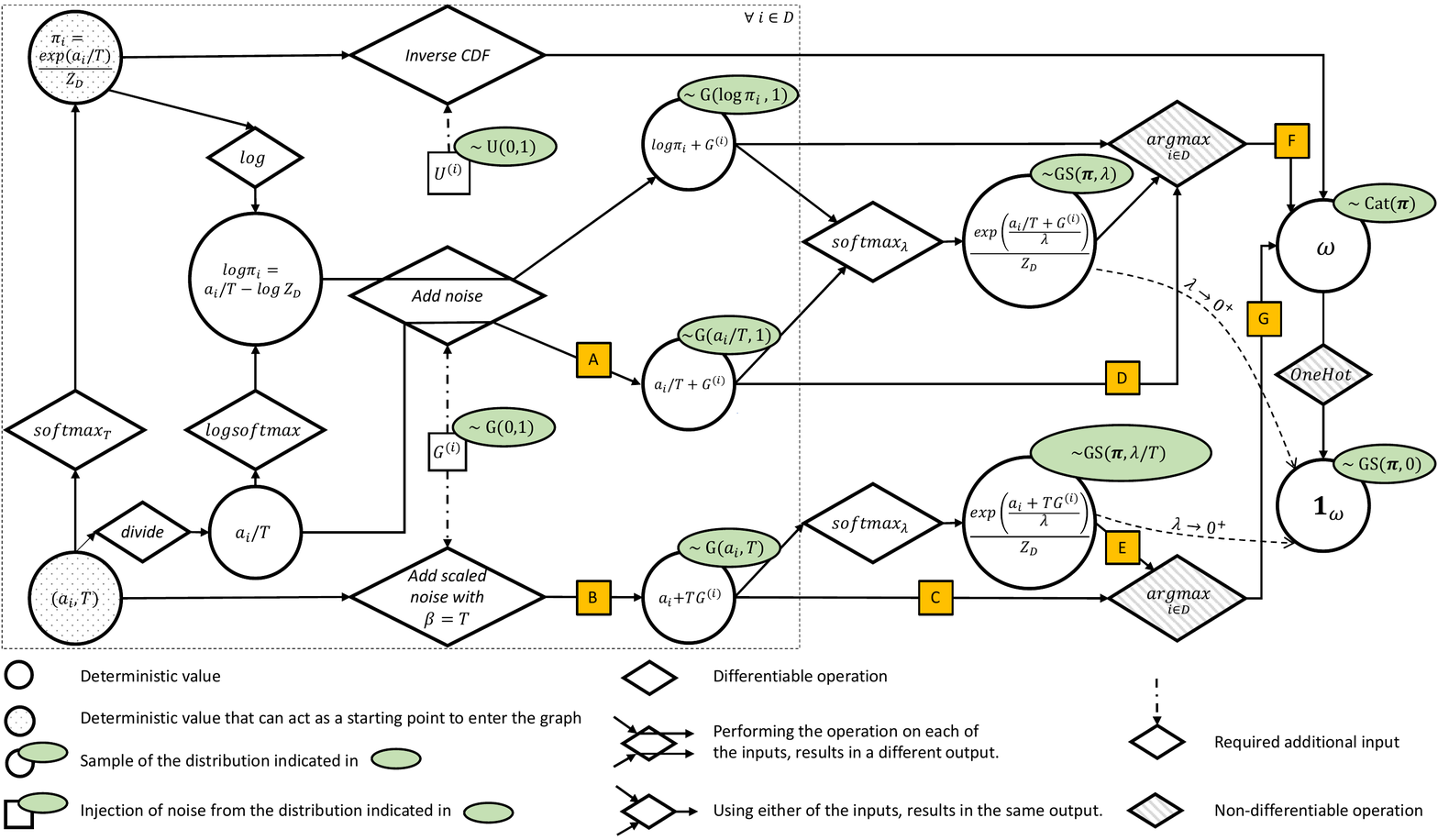}
    \caption[]{Drawing a sample from a categorical distribution can be done via different paths. The most suitable path depends on the context in which one would like to draw a sample. One can start from either the unnormalized log-probabilities $\va$ (and a Boltzmann temperature $T$), or the normalized probabilities $\boldsymbol{\pi}$. The green ellipses in the upper right corner of certain nodes indicate that the node represents a random variable following the distribution indicated in the ellipse. The different partition functions $Z_D$ all normalize their corresponding node, their input is omitted for readability reasons. The yellow square boxes are used to refer to certain paths in the text. A Jupyter Notebook that accompanies this figure is publicly available.\footnotemark}
    \label{fig:flat_draw_one_sample}
\end{figure*}

\subsubsection{Exploration vs exploitation}
\label{sec:boltzmann}

Sampling from a categorical distribution may in some applications adhere to an exploration-exploitation dilemma, where exploitation of already acquired information should be balanced with exploration of new information (e.g. when sampling an action in reinforcement learning). Pure exploitation often results in greedy solutions that appear sub-optimal in the long run, while continuously exploring new directions can also cause sub-optimality and instable or diverging behavior. 

A common approach to deal with this dilemma is Boltzmann exploration \cite{sutton1990integrated}, in which the temperature $T$ of the Boltzmann distribution is gradually lowered during training, therewith reducing the distribution's entropy, and moving from an exploration to an exploitation regime. Evidently, one can change the Boltzmann temperature and use inverse transform sampling to draw a sample from the categorical. On the other hand, from \cref{eq:scaled_gumbel_max} we see that samples from a tempered categorical distribution (i.e. $T \neq 1$) may also be drawn using the Gumbel-max trick, either by changing the Boltzmann temperature explicitly, or by changing the scale $\beta$ of the Gumbel distribution, from which independent samples are being drawn. Note from \cref{eq:gumbel_icdf} that such noise samples are simply scaled (with $\beta$) samples from the standard Gumbel distribution. Intuitively, down-scaling (i.e. setting $\beta < 1$) the injected noise in the Gumbel-max trick thus perturbs the logits in a lower extent, therefore giving rise to a lower-entropy categorical distribution from which we are effectively sampling. Figure \ref{fig:flat_draw_one_sample} visually relates sampling using an altered Boltzmann temperature (start in the bottom left and follow track \mbox{$A \veryshortarrow D \veryshortarrow F$}) and sampling using scaled Gumbel noise (start in the bottom left and follow track $B \veryshortarrow C \veryshortarrow G$). Figure \ref{fig:sampling_experiments} in Appendix \ref{app:sampling_experiments} shows results of sampling experiments for various values of $\beta$. 

This relation between the Boltzmann temperature and Gumbel noise scaling in the Gumbel-max trick, has been used by the authors of \cite{Cesa-Bianchi2017}, who propose Boltzmann–Gumbel exploration (BGE), an algorithm - inspired by the Gumbel-max trick - in which a class-dependent Gumbel noise scaling factor (i.e. $\beta(i)$) is used to guarantee sub-linear regret in a stochastic multi-armed bandit problem. In contrast to a class-\emph{independent} scaling factor, class-\emph{dependent} scaling factors do \emph{not} admit an analytical expression for the resultant categorical distribution. BGE was later leveraged in a recommender system setting by the authors of \cite{Biswas2019}, who more explicitly expressed the relation between the Boltzmann temperature and Gumbel noise scaling.

A similar exploration-exploitation trade-off exists in the field of natural text generation, where generated text should be of high quality (analogous to exploitation), but also diverse (analogous to exploration). Next to temperature scaling \cite{Shetty,Kojima2021}, other algorithms have been proposed to balance the aforementioned trade-off in this field, e.g. redistributing all mass of the Boltzmann distribution to either the $k$ classes covering the top-$k$ highest probabilities \cite{Fan2018b}, or the $p$ classes covering the top-$p$ ratio of the total mass \cite{Holtzman2019}. Note that for $\beta = 0$ in \cref{eq:scaled_gumbel_max} (i.e. no Gumbel noise is injected), the Gumbel-max trick resorts to setting $k=1$ in the work of \cite{Fan2018b}; i.e. all mass is redistributed to the class with the largest probability (resulting in a zero-entropy one-hot distribution) and sampling becomes a deterministic (argmax) function.

\begin{figure}
    \centering
    \includegraphics[width=1\textwidth,trim={0cm 5cm 3.2cm 1cm},clip]{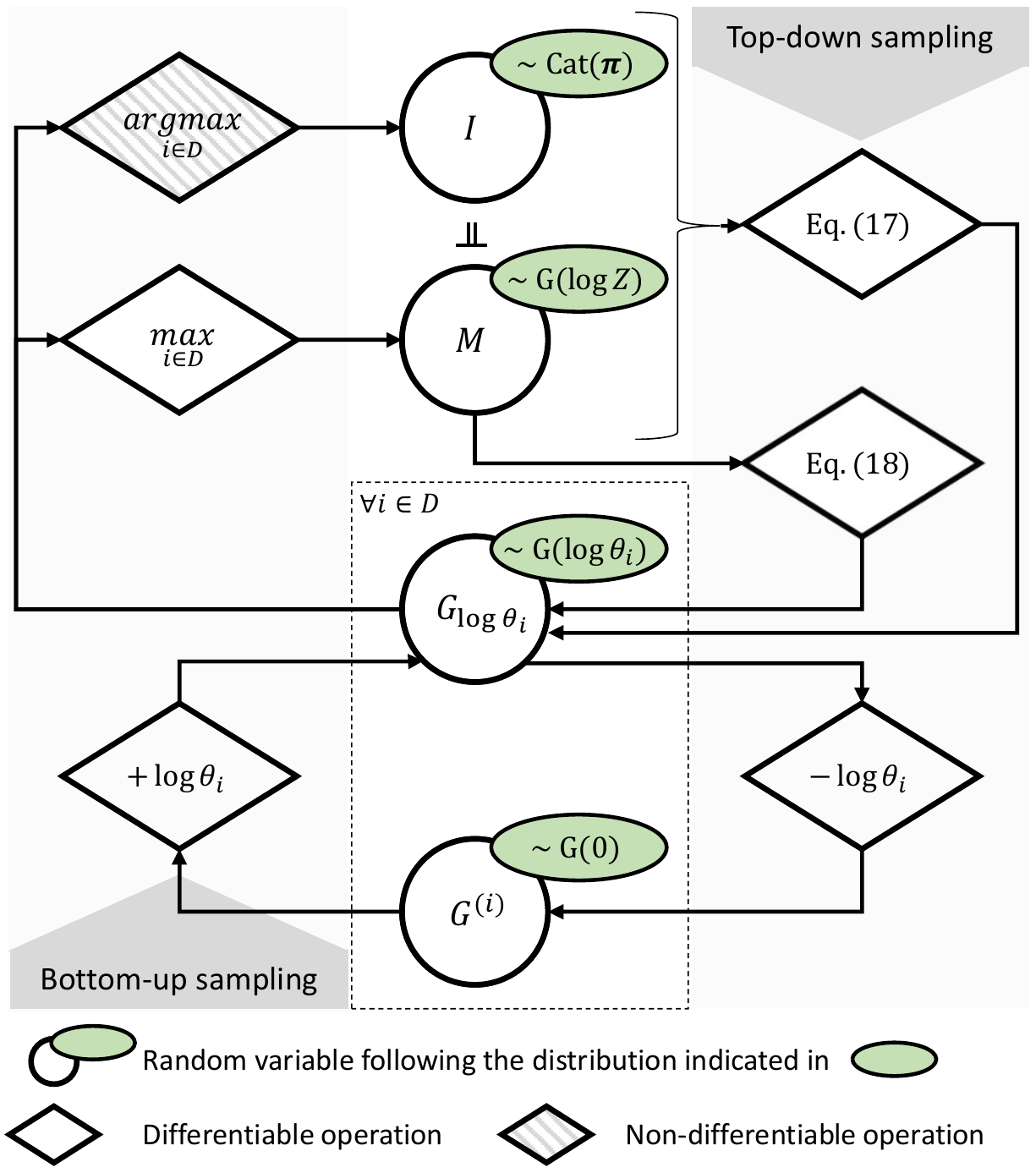}
    \caption{Sampling Gumbels and the corresponding argmax $I$ and maximum $M$, can be done in a bottom-up or top-down approach. The former finds the $\operatorname{(arg)max}$ from the perturbed logits, starting from the Gumbels. Top-down sampling, on the other hand, starts from either $M$ and/or $I$, and conditionally samples the corresponding perturbed logits. The maximum and argmax are independent ($\indep$), allowing for independent sampling of either of them in case only the other entity is known. The top-down procedure can be applied in parallel for the entire domain $D$ (depicted) or sequentially (not visualized).}
    \label{fig:bottom_up_top_down}
\end{figure}
\footnotetext{ \url{https://github.com/iamhuijben/gumbel_softmax_sampling}}
\subsection{Inverting the Gumbel-max trick: top-down sampling}
\label{sec:gumbel_max_revert}

Both the argmax and max function in \cref{eq:gumbel_max} and \cref{eq:max_stability}, respectively, are surjective/many-to-one mappings, i.e. numerous (multi)sets\footnote{A multiset or bag is a set in which an element can occur multiple times.} of perturbed logits exist that result in the same index $I$ or maximum $M$. In certain scenarios one can be interested in reverting this sampling process, i.e. inferring the shifted Gumbels (i.e. perturbed logits) that may have generated a particular sample/index or maximum. This reversion is often referred to as \emph{top-down sampling} \cite{Maddison2014a,Kool2019a}. To denote the standard procedure, i.e. the `conventional' Gumbel-max trick in which Gumbels are sampled unconditionally, the term \emph{bottom-up} sampling was introduced \cite{Maddison2014a}. The notion of top-down sampling gave rise to several applications, ranging from Gumbel-based sampling algorithms \cite{Maddison2014a, Kool2019a, Kool2020a,Shi2020, Wiggers2020}, gradient estimators \cite{Tucker2017,Grathwohl2018,Paulus2021}, counterfactual outcome prediction in structural causal models \cite{Oberst2019}, and discrete flow models \cite{hoogeboom2021}. We will elaborate on the principles of top-down sampling in this section.

We seek a set of $N$ independently perturbed logits, with a maximum value $M=m$, located at index $I=\omega$. Given either of the two (i.e. the maximum or index), we can - thanks to their independence - always sample the other from $\operatorname{Cat}(\boldsymbol{\pi})$ (sampling $I$) or $\operatorname{Gumbel}(\log Z)$ (sampling $M$). Once both the maximum and corresponding index are known/sampled, the value of all ($N-1$) other perturbed logits should now be restricted to be $\leq m$. Acquiring Gumbels, given this restriction, can be achieved by drawing them from a right-truncated Gumbel distribution \cite{Maddison2014a,maddison2017_blog}:
\begin{equation}
    \tilde{G}_{\log\theta_i1,m} \sim \operatorname{TruncGumbel}(\log \theta_i,1,m),
\end{equation}

\noindent with the ICDF of $\operatorname{TruncGumbel}(\mu,\beta,m)$ being:
\begin{equation}
    \label{eq:icdf_trunc_gumbel}
    F^{-1}(u) = -\beta\log(\exp(\mu-\frac{m}{\beta})-\log u) + \mu.
\end{equation}

\noindent Analogous to the max-stability property given in \cref{eq:max_stability}, the maximum of a set of independent samples from $\operatorname{TruncGumbel}(\log \theta_i,1,m)$, follows again a truncated Gumbel distribution:
\begin{equation}
    \label{eq:gumbel_max_trunc}
        \underset{i\in D}{\operatorname{max}}\{{\tilde{G}_{\log \theta_i,1,m}}\} \sim \operatorname{TruncGumbel}(\log Z,1,m).
\end{equation}

\noindent The top-down sampling procedure can be summarized as follows, $\forall i \in D$:
\begin{align}
\label{eq:conditional_gumbels}
G_{\log \theta_i,1}|\{\omega,m\} = \begin{cases}
m  &\hspace{0.2cm} i = \omega\\
    -\log(\exp(-m)-\frac{\log U^{(i)}}{\theta_i}) &\hspace{0.2cm} i \neq \omega.
\end{cases}
\end{align}

\noindent The second case ($i \neq \omega$) expresses the ICDF of $\operatorname{TruncGumbel}(\log \theta_{i},1,m)$. In case $m$ was sampled from $\operatorname{Gumbel}(\log Z)$ (rather than given), \mbox{$m = -\log(-\log U^{(\omega)}) + \log Z$}, and $\exp(-m)$ reduces to $-\log(U^{(\omega)})/Z$. Equivalent expressions were given by the authors of \cite[App. C]{Tucker2017} (assuming $Z=1$), and \cite[Eq. (9)]{Paulus2021}.

The authors of \cite{Kool2019a} observe that, when conditioning on $M=m$, we can even directly provide an expression for all perturbed logits without explicitly sampling $I=\omega$ first. More specifically, a set of (unconditional) Gumbels $G_{\log \theta_i} \sim \operatorname{Gumbel}(\log \theta_i)~\forall i \in D$, with maximum $q = \underset{i\in D}{\operatorname{max}}\{G_{\log \theta_i}\}$, can be transformed to conditional Gumbels $\tilde{G}_{\log \theta_i,1,m}$, via the respective (inverse) CDFs. More specifically, $\forall i \in D$:
\begin{equation}
    \label{eq:transform_to_trunc_gumbels}
    \tilde{G}_{\log \theta_i,1,m} := G_{\log \theta_i,1}|m = F_{m}^{-1}(F_q(G_{\log \theta_i})),
\end{equation}
where $F_q(\cdot)$ and $F_{m}^{-1}(\cdot)$ denote the CDF and ICDF of truncated Gumbel distributions\footnote{The ICDF of $\operatorname{TruncGumbel}(\mu,\beta,m)$ is provided in \cref{eq:icdf_trunc_gumbel}.\\The CDF is expressed as \\\mbox{$F(x) = \exp(-\exp(-\frac{\operatorname{min}(x, m)-\mu}{\beta}))/ \exp(-\exp(-\frac{m-\mu}{\beta}))$}.}, both located at $\log \theta_i$, and truncated at $q$ and $m$, respectively. The set $\{\tilde{G}_{\log \theta_i,1,m}\}_{i\in D}$ is now ensured to have a maximum $m$. Figure \ref{fig:bottom_up_top_down} visually summarizes both the bottom-up and the top-down sampling procedure.

Rather than simultaneously sampling all conditional Gumbels for $i\neq \omega$, they could also be acquired sequentially. This enables conditional Gumbel sampling for applications in which one is not interested in the perturbed logits over the entire domain $D$, because it is typically too large. The authors of \cite{Maddison2014a} note that hybrid top-down sampling formats (i.e. approaches that are in between simultaneous and fully sequential sampling) can also be used without loss of generality. They propose to partition the sampling space into two (random and mutually exclusive) sub-domains, and apply the sequential process within these domains separately. Their top-down construction algorithm generalizes the sequential conditional sampling procedure, and is provided in Appendix \ref{app:top_down}.

\subsection{Extended sampling algorithms}
\label{sec:extended_sampling_algorithms}

\begin{figure*} 
    \centering
    \includegraphics[width=\linewidth,trim={0cm 3.cm 0cm 1cm},clip]{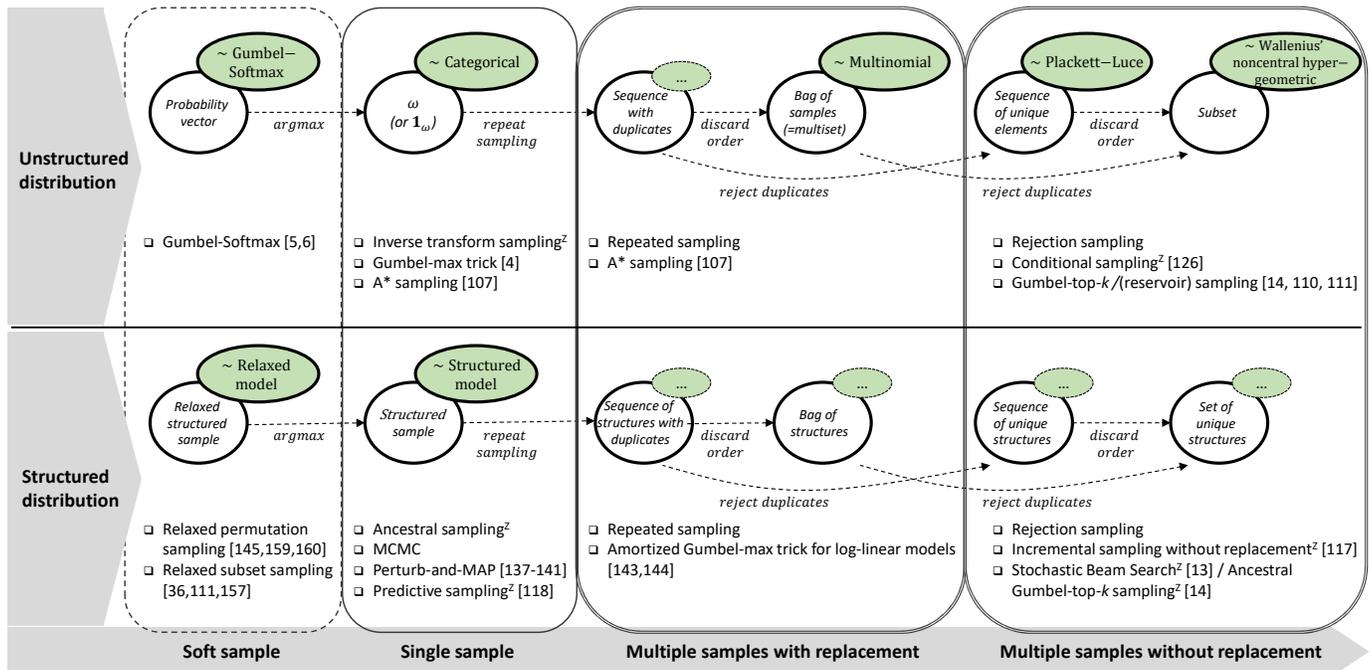}
    \caption{Overview of sampling algorithms for different scenarios: single (discrete or soft) or multiple sample(s) from (un)structured distributions. For all cases, both default/conventional and Gumbel-based algorithms are given. The green ellipses indicate the distribution from which samples of the indicated algorithms originate. In case of empty ellipses, it is unknown which distribution the samples follow. A superscript $z$ indicates that the algorithm requires normalized probabilities, i.e. partition function $Z$ should be known.}
    \label{tab:sampling}
\end{figure*}

The Gumbel-max trick requires $N$ Gumbel realizations to draw one sample from a categorical distribution of $N$ classes. This can become cumbersome when drawing many samples, or even intractable when drawing a sample from an exponentially large domain (i.e. large $N$). Fortunately, its theoretical properties naturally extend to sampling algorithms with varying purposes, e.g. sampling with or without replacement, and sampling from structured models. This section discusses such Gumbel-based sampling algorithms, which are summarized in \cref{tab:sampling}, including conventional alternatives. The top and bottom row display the unstructured and structured setting, as discussed in \cref{sec:unstructured_sampling} and \ref{sec:structured_sampling}, respectively. Algorithms that require a normalized distribution, i.e. a known partitioning function $Z$, are denoted with a superscript $z$. Green ellipses indicate the distribution (when known), from which algorithms return samples. Note that soft samples (first column of \cref{tab:sampling}) are discussed in \cref{sec:gradient_estimation_NN}.

\subsubsection{Unstructured distributions}
\label{sec:unstructured_sampling}

Drawing a single sample from an (unstructured) categorical distribution can among others be done using inverse transform sampling (\cref{sec:standard_sampling}), and the Gumbel-max trick (\cref{sec:gumbel_max}).  Figure \ref{fig:flat_draw_one_sample} visually relates these sampling algorithms. The $A^*$ sampling algorithm \cite{Maddison2014a} can be interpreted as a continuous counterpart of the Gumbel-max trick, that samples from continuous domains (i.e. $N \rightarrow \infty$). $A^*$ sampling does - analogously to the Gumbel-max trick - not require knowledge about the partition function $Z$. It leverages top-down sampling (as explained in \cref{sec:gumbel_max_revert}) and bounds related to Gumbel processes \cite{Maddison2014a}, to diminish the number of i.i.d. Gumbels to be drawn, while still guaranteeing exact samples from unnormalized continuous distributions.\\

\noindent \textbf{Sampling with replacement\\}
Repeating a discrete sampling process, we can draw multiple (independent) samples \emph{with} replacement. Depending on whether the order of samples is considered, this gives rise to an (ordered) \emph{sequence}, or an (unordered) \emph{bag} or \emph{multiset}
of samples. Such sequences and bags (or certain representations thereof) can in itself be considered samples from structured models (see Section \ref{sec:structured_sampling}). For example, when a bag of samples is represented as a vector of counts over the complete categorical sampling domain, this vector has a (structured) distribution that is known as the multinomial distribution. Formally, the counts-vector resulting from drawing $k$ samples (with replacement) from $\operatorname{Cat}(\boldsymbol{\pi})$ is a \emph{single} sample from $\operatorname{Multinomial}(\boldsymbol{\pi}, k)$. Therefore, the same sampling process may give rise to different distributions, depending on what is considered `the sample'. 

When repeated samples from the categorical are drawn using the Gumbel-max trick, it may be possible to gain efficiency by reusing computations, as proposed by the authors of \cite{Maddison2014a,Qi2020}. \\


\noindent\textbf{Sampling without replacement\\}
\label{sec:sampling_wo_replacement}
When \emph{rejecting} (or removing) duplicates from the bag or sequence of samples drawn with replacement, we obtain a set, respectively sequence, of \emph{unique samples}. This can be seen as \emph{sampling without replacement}, implemented using rejection sampling, where proposals (i.e. individual samples from the bag/sequence) get rejected if they have already been sampled. Depending on the distribution, it may require many proposal samples to acquire a sequence or set of $k$ unique elements (e.g. repeated sampling from a low-entropy distribution often returns the same class). Therefore, a more common alternative to rejection sampling is to \emph{sequentially sample without replacement} by removing a sampled element from the sampling domain, renormalize the distribution, and continue to draw the next (unique) sample from the updated domain. The subsequent (conditional) samples are typically sampled using standard inverse transform sampling.

If the order of the resulting unique samples is considered, this process gives rise to a probability distribution over sequences $X$ of length $k$:
\begin{equation}
\label{eq:p_ordered_wor}
    p(X) = \prod_{i=1}^k \frac{p(x_i)}{1 - \sum\limits_{j<i} p(x_j)}.
\end{equation}
When $k = N$, i.e. the full domain is sampled, this represents a distribution over \emph{permutations} known as the Plackett-Luce model \cite{Plackett1975,luce1959individual}. When we do not care about the order of the unique samples, we can find the probability for a certain set of samples $\mathcal{S}$, by summing the probabilities for all possible permutations of $\mathcal{S}$:
\begin{equation}
\label{eq:p_unordered_wor}
    p(\mathcal{S}) = \sum_{X \in~\text{Perm}(\mathcal{S})} p(X). 
\end{equation}
This probability distribution is a special case of the Wallenius' noncentral hypergeometric distribution \cite{wallenius1963biased,chesson1976non}, and can be computed exactly in $O(N!)$, or approximated by an integral \cite{chesson1976non, fog2008calculation}.

\begin{figure}
\centering
\includegraphics[width=0.6\linewidth,trim={0.5cm 12.5cm 19.5cm 1.8cm},clip]{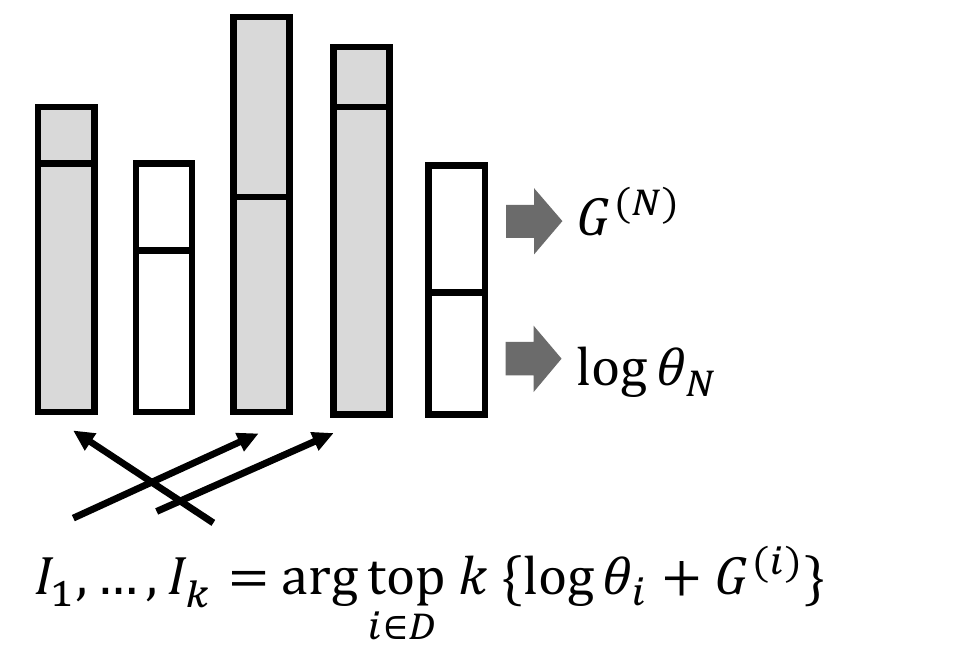}
\caption{Illustration of Gumbel-top-$k$ sampling. Figure adapted from \cite{Kool2020a}. Gumbel-top-$k$ perturbs all logits with i.i.d. Gumbel samples, and returns the index of the $k$ perturbed logits with the highest values. These indices yield $k$ independent samples without replacement from $\operatorname{Cat}(\boldsymbol{\theta})$.}
\label{fig:gumbel_topk}
\end{figure}

Renormalizing the distribution for sequential sampling without replacement (as done in \cref{eq:p_ordered_wor}) may be difficult/intractable, especially in large domains. An appealing alternative is to use the Gumbel-max trick on the updated domain (being a subset of the original sampling domain), as it avoids the need for explicit renormalization. Even more interesting, when we apply the Gumbel-max trick repeatedly for sampling without replacement, the $N$ perturbed logits can be \emph{reused} for all $k$ samples, by simply selecting the top-$k$ perturbed logits in a single step \cite{vieira2014gumbel, Xie2019}, see \cref{fig:gumbel_topk}. This algorithm, which has become known as \mbox{Gumbel-top-$k$} sampling \cite{Kool2020a}, is a strict generalization of the Gumbel-max trick (which is the special case for $k = 1$). This parameterization of sampling without replacement allows for another derivation of the integral to approximate \cref{eq:p_unordered_wor}, or to compute it exactly in $O(2^N)$ \cite{Kool2020b}.

Gumbel-top-$k$ sampling is analogous to weighted reservoir sampling, as shown by \cite{vieira2014gumbel,Xie2019}. The latter uses the Beta-distribution parameterization (see \cref{sec:uni_exp_gumbel_section}), and selects the top-$k$ perturbed keys using a priority queue, enabling sampling from streaming applications \cite{Efraimidis2006}. Gumbel-top-$k$ sampling also has connections to ranking literature, as it was shown by \cite{Yellott1977} that the Gumbel distribution is the \emph{only} distribution of discriminal scores in a Thurstonian ranking model \cite{thurstone1927}, inducing a ranking that satisfies Luce's axiom of choice \cite{luce1959individual}. 

Samples without replacement can be used to form statistical estimators of functions of the underlying categorical distribution \cite{raj1956some,murthy1957ordered,duffield2007priority,vieira2017estimating,Kool2020b}. Similar ideas have been used to construct gradient estimators for the process of sampling without replacement (see \cref{sec:gradient_estimators}).

\subsubsection{Structured distributions}
\label{sec:structured_sampling}

The Gumbel-max trick can be seen as an instance of perturb-and-MAP \cite{Papandreou2011}, which is a class of methods that transform sampling into an optimization problem, where the sample corresponds to the optimum of a perturbed energy function. Specifically, in the context of the Gumbel-max trick, the energies correspond to (unnormalized) log-probabilities and the argmax operator represents the optimization problem, i.e. `finding the maximum' of the perturbed log-probabilities/energies. Whereas finding the maximum in an unstructured setting with a limited number of classes is hardly considered a `problem', it is challenging in structured models, where there are multiple dependent variables with a sampling domain of exponential size. Various (approximate) sampling algorithms have been proposed, which rely on the perturb-and-MAP principle by finding the (approximate) maximum of the perturbed energy over all possible configurations of such structured models \cite{Papandreou2011,Tarlow2012,Hazan2013,Chen2016,Kim2016}.

Note that there is a vast amount of literature on alternatives to sample from structured models. Standard \emph{ancestral sampling} can, for example, be used when the conditional distributions for variables are normalized (i.e.\ the partition function is known), whereas Markov Chain Monte Carlo (MCMC) sampling \cite{hastings1970monte} and variants thereof can be leveraged when $Z$ is unknown. In this section, we focus on methods related to Gumbel variables, such as the perturb-and-MAP paradigm.

The Gumbel-max trick can be seen as a \emph{reparameterization trick} \cite{kingma2013,rezende2014}, that allows to sample a categorical variable as a deterministic transformation of independent (Gumbel) noise variables. This reparameterization property is exploited by the authors of \cite{Wiggers2020},  that aim to parallelize sampling by forecasting (reparameterized) samples from a sequential model.\\ 

\noindent \textbf{Multiple samples\\}
Whereas samples with or without replacement (from unstructured distributions) can in itself be seen as a single sample from a structured distribution (e.g. over sets or sequences), it is also possible to draw multiple samples (with or without replacement) from structured distributions themselves (see bottom right corner of \cref{tab:sampling}). For example, we can draw a set (or sequence) of (unique) sequences from a (neural) sequence model. The authors of \cite{Mussmann2016,Mussmann2017} show how to efficiently compute multiple (structured) samples with replacement from unnormalized distributions using the perturb-and-MAP paradigm. When a structured distribution is normalized, such as in neural sequence models, generalizations of ancestral sampling can be used to efficiently draw samples without replacement. Examples of such algorithms are \emph{stochastic beam search} \cite{Kool2019a}, or its generalization \emph{ancestral Gumbel-top-$k$ sampling} \cite{Kool2020a}, which implicitly apply Gumbel-top-$k$ sampling using top-down sampling (\cref{sec:gumbel_max_revert}), and \emph{UniqueRandomizer} \cite{Shi2020}, which applies sequential sampling without replacement using an efficient data structure to compute the necessary renormalizations. 

\section{Gradient estimation in neural network optimization}
\label{sec:gradient_estimation_NN}

Section \ref{sec:applications} introduced various applications of gradient estimation of discrete stochastic nodes in neural network optimization. Most Gumbel-based estimators (as discussed in \cref{sec:gradient_estimators}) build upon the recently introduced Gumbel-Softmax distribution, which will be discussed in \cref{sec:gumbel_softmax}. 

\subsection{Gumbel-Softmax distribution}
\label{sec:gumbel_softmax}
Instead of drawing hard/discrete samples from an (unstructured) categorical distribution, one can also define soft samples, which may notably be used for gradient estimation. To see the relation between these hard and soft samples, we must consider the hard samples in their one-hot embedding format, i.e. $\mathbbm{1}_{\omega} \in \{0,1\}^N$. A soft sample $S_\lambda$ can then be defined as a vector of equivalent length, in which the mass is `spread out' over multiple bins, rather than centered in one class. Concurrent works \cite{Jang2017,Maddison2017} have introduced the Gumbel-Softmax (GS; or Concrete) distribution, of which an exact sample is a relaxation of $\mathbbm{1}_{\omega}$. We refer to \cite{Jang2017,Maddison2017} for (derivation of) the PDF of this distribution, here denoted with $\operatorname{GS}(\boldsymbol{\pi},\lambda)$. More specifically, the $i^{\mathrm{th}}$ index of soft sample $S_\lambda \in \{\mathbb{R}^N_{\geq 0}: |S_\lambda|=1\}$ is defined as:

\begin{equation}
    \label{eq:gumbel_softmax}
    S_{i;\lambda} = \frac{\exp\big((\log \theta_i+G^{(i)})/\lambda\big)}{\sum_{j\in D} \exp \big((\log \theta_j + G^{(j)})/\lambda \big)},
\end{equation}
where $\lambda$ is a temperature parameter that influences the entropy of both the GS distribution, and the corresponding samples. Viewing these samples as relaxations of samples from $\operatorname{Cat}(\boldsymbol{\pi})$, $\lambda$ can also be interpreted as a parameter that defines the `amount of relaxation' of soft sample $S_\lambda$. In the limit of $\lambda \rightarrow 0^{+}$, $S_\lambda$ converges to hard sample $\mathbbm{1}_{\omega}$, and the PDF of the GS distribution converges to $\operatorname{Cat}(\boldsymbol{\pi})$. In fact, drawing a GS sample can be interpreted as solving an entropy-regularized (by $\lambda$) linear program on the probability simplex \cite{Mena2018}.

In the special case of $N=2$, i.e. when the categorical distribution converges to the Bernoulli distribution, the softmax function in the GS distribution can be replaced by its binary counterpart, the sigmoid function \cite{Jang2017,Maddison2017}.

Analogously to the Gumbel-max trick, $S_\lambda$ is invariant to scaling of the unnormalized probabilities in $\boldsymbol{\theta}$: scaling (the entire vector with the same constant) in probability space results in addition in log-space, which leaves the translation-invariant softmax function in \cref{eq:gumbel_softmax} unaffected. Moreover, relaxing $\mathbbm{1}_{\omega}$, using $S_\lambda$, preserves order, i.e. $S_\lambda$ can immediately be transformed to $\mathbbm{1}_{\omega}$ via the argmax operation (see the upper left cell in \cref{tab:sampling}). Figure \ref{fig:flat_draw_one_sample} visually relates hard and soft samples. Also the relation between both temperatures $\lambda$ and $T$ (see track $B \veryshortarrow E$) is shown, which will further be discussed in \cref{sec:T_and_lambda}. Figure \ref{fig:sampling_experiments}  in Appendix \ref{app:sampling_experiments} shows results of sampling experiments with various values of $\lambda$. An interactive setup to repeat these experiments is provided with this work\footnote{\url{https://iamhuijben.github.io/gumbel_softmax_sampling.html}}.

Although the GS distribution has (so far) found most use as a means to relax discrete stochastic nodes for differentiable neural network optimization, it has recently also found use in relaxing a maximum-a-posteriori (MAP) objective in the context of inverse problems with discrete multi-variate random variables \cite{Beck2020}, and in solving combinatorial problems \cite{Li2020a}.

\subsection{Gradient estimators}
\label{sec:gradient_estimators}

Most of the applications discussed in Section \ref{sec:applications} pose modular machine learning models that involve one or several discrete stochastic components. We highlight the challenges posed by training such models by considering a simplified setting with the following optimization program:
\begin{equation}
\label{eq:nn_objective}
\min_{\phi} \E_X[f(X)] \, \quad \text{ with } X \sim p_{\phi},
\end{equation}
where $X$ is a random variable, $p_\phi$ is a probability distribution parameterized by $\phi$, and $f$ some function with range in $\mathbb{R}$. The expectation is taken over random variable $X$ and notably depends on $\phi$. Using gradient-based methods to optimize \cref{eq:nn_objective} requires an estimator\footnote{If $p_{\theta}$ is designed to be a degenerate distribution, i.e. there is no stochasticity (e.g. as in \cite{van2017neural}), both the expectation and its gradient do not need to be estimated. Therefore, we only consider cases where $p_{\theta}$ is a non-degenerate distribution.} for $\nabla_{\phi} := d\E_X[f(X)] / d\phi$. There are at least two groups of estimators that could be considered, \emph{score-function estimators} and \emph{pathwise estimators} \cite{mohamed2020monte}.

The score-function estimator, also known as REINFORCE estimator \cite{glynn1990likelihood, williams1992simple}, is in its simplest form given by
\begin{equation}
\label{eq:reinforce}
	\nabla_{\text{REINF}} := f(X)\frac{\partial \log p_\phi(X)}{\partial \phi}.
\end{equation}
This estimator makes minimal assumptions on $p_{\phi}$ and $f$ and is notably applicable when $X$ is a discrete random variable. The estimator is unbiased \cite[Section 4.3]{mohamed2020monte}, but tends to have high variance that may render it ineffective in practical applications. To reduce its variance, numerous gradient estimators based on \cref{eq:reinforce} have been proposed. Many of these estimators involve control variates \cite{Tokui2017, Tucker2017, Grathwohl2018, mnih2014neural}, sometimes in conjunction with drawing multiple samples from the discrete distribution \cite{mnih2016variational, Kool2019b, Kool2020b, Yin2019a}.

Pathwise estimators  are also known as reparameterization gradient estimators \cite{Salimans_2013, kingma2013}. They reparameterize the random variable $X$ using an auxiliary random variable $\epsilon$ and a deterministic path $h(\epsilon, \phi)$, such that $X = h(\epsilon, \phi)$ for $\epsilon \sim p(\epsilon)$. This reparameterization can be exploited to interchange the differential and expectation operators, giving rise to a gradient estimator of the form
\begin{equation}
\label{eq:reparam}
	\nabla_{\text{REPARAM}} := \frac{\partial f(X)}{\partial X}
	\frac{\partial h(\epsilon, \phi)}{\partial \phi}.
\end{equation}
This estimator makes more assumptions on $p_{\phi}$ and $f$ \cite[Section 5]{mohamed2020monte}. It, for example, requires $f$ to be differentiable and the reparameterization of $X$ to be continuous. When well-defined, the pathwise estimator is unbiased and tends to have low variance, which makes it a popular choice in practice.  Unfortunately, discrete random variables do not admit reparameterization gradient estimators, because their reparameterizations are discontinuous (e.g. consider the discontinuous argmax operator in the Gumbel-max trick). Therefore, for discrete random variables, relaxed gradient estimation was proposed \cite{Jang2017, Maddison2017, Paulus2020}. Relaxed gradient estimators approximate a discrete random variable with a continuous random variable that admits a reparameterization gradient. This reparameterization gradient is then used as a biased gradient estimator for the parameters of the discrete distribution. Notably, the continuous random variable is only used during training to give a gradient estimator, while at test time the model is evaluated using the discrete random variable. For various discrete distributions, relaxed gradient estimators have been proposed, which we now briefly survey.  

\subsubsection{Gradients for sampling unstructured distributions}
When $X$ is a categorical random variable, the GS estimator \cite{Jang2017, Maddison2017} is a popular estimator. It approximates the categorical distribution with the GS distribution (as presented in \cref{sec:gumbel_softmax}). The estimator is given by 
\begin{equation}
\label{eq:gs_estimator}
	\nabla_{\text{GS}} := \frac{\partial f(S_{\lambda})}{\partial S_{\lambda}}\frac{\partial S_{\lambda}}{\partial \phi},
\end{equation}
where $S_\lambda = \operatorname{softmax}_{\lambda}\left(\phi + \boldsymbol{G} \right)$, with $\phi = \log \boldsymbol{\theta}$ and $\boldsymbol{G}$ a vector of i.i.d. Gumbel variables (see \cref{eq:gumbel_softmax}).
For this estimator to be well-defined, $f$ must be defined and differentiable on the interior of the simplex. The GS estimator reduces variance at the expense of introducing bias: $\nabla_{\text{GS}}$ is an unbiased estimator of $d\E_{S_{\lambda}}\left[f(S_{\lambda})\right]/d\phi$, but a biased estimator of $\nabla_{\phi}$. Empirically, the temperature parameter $\lambda$ of the GS distribution can be used to trade-off bias and variance of the gradient estimator. Lower temperatures tend to be associated with higher variance, but reduced bias. An important variant of the above estimator is the straight-through GS (ST-GS) estimator \cite{Jang2017, Paulus2021} that replaces $S_{\lambda}$ with $X$ in the first term of \cref{eq:gs_estimator} (i.e. it uses non-relaxed computations in the forward pass). Other work has proposed the use of alternative continuous distributions to obtain relaxed gradient estimators for categorical distributions \cite{Potapczynski2019, Paulus2020} that may lead to improvements in some applications. 

\subsubsection{Gradients for sampling structured distributions}
Recent works have extended the GS estimator from categorical distributions to structured distributions \cite[and references therein]{Paulus2020}. For example, various gradient estimators have been proposed for the case where $X$ is a subset of fixed size \cite{Chen2018a, Xie2019, Paulus2020}. The authors of \cite{Chen2018a} propose a relaxation that uses repeated samples from the GS distribution to relax subset selection, and \cite{Xie2019} use successive applications of the softmax function. The authors of \cite{Paulus2020} present a framework that naturally extends the Gumbel-max trick and its relaxation to other structured variables and use it to define various relaxed gradient estimators for subset selection. For the case where $X$ is a permutation, the GS distribution has been generalized into the Gumbel-Sinkhorn distribution to obtain a relaxed gradient estimator \cite{Mena2018}, and the authors of \cite{Linderman2018} and \cite{Grover2019} develop gradient estimators based on alternative transformations.

\section{Practical considerations}
\label{sec:considerations}

When using Gumbel-based (soft) sampling algorithms, several implementations and settings might be considered. This section reviews distinct options and commonly-made choices.

\subsection{Parameterization of the logits}

Each of the presented algorithms in \cref{sec:sampling_algorithms} (see \cref{tab:sampling}) rely on availability of (normalized or unnormalized) probabilities of a categorical distribution. These probabilities can either be considered trainable parameters (i.e. they are optimized over the full training set), yielding direct optimization, or they can be optimized indirectly, e.g. by predicting them by a neural network. Both options will be discussed below.

\subsubsection{Direct parameterization}
When considering direct optimization of the unnormalized log-probabilities (or logits), the number of trainable parameters typically equals the number of classes in case of a flat/unstructured distribution. For structured distributions, the number of trainable parameters is typically smaller than the number of possible structures. The work of \cite{Huijben2020a}, e.g., distinguishes two parameterizations for subset sampling. On the one hand $k$ samples without replacement can be drawn from one set of logits (referred to as top$K$ sampling in their work), while on the other hand these samples could each be drawn from their own categorical distribution (top$1$ sampling). Even though the top$1$ parameterization allows for learning dedicated distributions for each of the samples, this parameterization was not found to consistently outperform the top$K$ counterpart. A similar observation was reported by \cite{Baevski2020}, where on par results were found when either sharing or not sharing learned (via ST-GS) quantization codebooks among latent groups. Imperfect optimization could explain these observations.

\subsubsection{Indirect parameterization}
Logits are often parameterized by a neural network when conditionality on incoming data is required. Example applications of such data-conditionality entail NAS \cite{Veit2017,Bejnordi2019, Wang2020a}, attention \cite{Kang2019}, active data acquisition \cite{vanGorp2021}, and when GS is used at the output of a GAN's generator \cite{Kusner2016}. The number of trainable parameters that parameterize the logits now depends on the architecture of the logit-predicting model. Despite being unrestricted in size, such models are often designed to be simple and shallow in practice \cite{Shu2018,Chen2018,Kang2019,Bejnordi2019, Baevski2020,vanGorp2021}.

Taking a different approach, the GS estimator can also be used without explicitly predicting the logits themselves, but by inferring them from already drawn samples from the categorical distribution \cite{Gu2018}, leveraging conditional Gumbel sampling \cite{Maddison2014a} (see \cref{sec:gumbel_max_revert}).

\subsubsection{Standard implementations}
Independent of the logits being directly or indirectly parameterized, once the unnormalized logits are known and one is interested in drawing one GS-sample from a set of unnormalized logits, implementations from popular deep learning libraries Tensorflow\footnote{\url{https://www.tensorflow.org/agents/api_docs/python/tf_agents/distributions/gumbel_softmax/GumbelSoftmax}}~\cite{tensorflow} and Pytorch\footnote{\url{https://pytorch.org/docs/stable/generated/torch.nn.functional.gumbel_softmax.html}}~\cite{pytorch} can be used. The Pytorch implementation also facilitates the ST-GS estimator, i.e. drawing a hard sample in the forward pass, which is not supported by the current Tensorflow implementation. More extended Gumbel-based sampling algorithms and corresponding gradient estimators have not yet been implemented in standard libraries.

\subsection{Initialization and regularization}
Considerations can be made regarding the initialization and regularization of the logits in the Gumbel-(soft)max trick. Typical initializations consider small random values \cite{Abid2019}, e.g. drawn from a Uniform \cite{Guo2020} or Gaussian distribution \cite{Huijben2020a}, or set all logits to zero \cite{Lacey2018}. Others have based initialization on \textit{a priori} knowledge regarding the distribution \cite{Chen2018,Huijben2019,Yang2020c}, or a desired sampling rate \cite{Veit2017,Louizos2018b,Schlichtkrull2021}.

Also various regularization strategies have been proposed. In the context of model pruning, logits often represent the (unnormalized) log-probability with which certain elements (e.g. layers, kernels etc.) should be selected (i.e. sampled) to be part of the model architecture. In this field, a sparsity-promoting loss is often leveraged to penalize dense and/or large models, which indirectly influences the logits during training \cite{Bejnordi2019,Kang2020,Wan2020,Herrmann2020,Verelst2020,Schlichtkrull2021,Cai2021}. 

With the goal of learning a subsampling matrix for data compression, the authors of \cite{Huijben2020a} regularize the logits more directly by promoting low-entropy distributions. Conversely, when GS is used in discrete GANs, entropy of the generator's (output) distribution is often maximized rather than minimized, in order to stabilize optimization and prevent mode collapse (which is the case when the generator outputs non-diverse samples) \cite{Gu2018,Huang2019,Bragman2019}. To deal with this diversity-quality dilemma in GANs, also distance regularization in latent space has been leveraged \cite{Shetty,Zhao2020}, promoting closeness of embeddings, and therefore (indirectly) influencing the logits predicted by the generator.

Finally, regularization has also been based upon \textit{a priori} knowledge, e.g. about a prior distribution, for which the KL divergence between the learned logits and this prior can be leveraged \cite{Havrylov,Gal2017}, or about labels in semi-supervised learning \cite{Tjandra2019,Fu2020a}. In the latter setup, the authors of \cite{Tjandra2019,Fu2020a} penalize the categorical cross-entropy loss on the predicted probabilities in a latent space, given some labels. 

\subsection{Setting the Gumbel-Softmax temperature \texorpdfstring{$\lambda$}{lambda}}
As introduced in \cref{sec:gumbel_softmax}, the GS distribution introduces an additional hyper-parameter $\lambda$. The authors of \cite{Maddison2017} show that, as long as $\lambda \leq (N-1)^{-1}$, with $N$ being the number of classes, the GS distribution is log-convex. In other words, it will not have a mode in the interior of the $N$-$1$-simplex. This is a desirable property of a relaxation, in order to not relax discreteness `too much', which could result in a considerable performance gap when subsequently evaluating the (non-relaxed) discrete model. Despite this theoretical insight, the empirically optimal value was, nevertheless, often found to be higher \cite{Maddison2017}. This may be a result of the bias-variance trade-off which is present when gradient estimation is done on GS samples. A high value of $\lambda$ induces high bias, but low variance, while a lower value diminishes this bias, but increases the variance, and leads to vanishing gradients for $\tau \rightarrow 0^+$. In line with \cite{Maddison2017}, the authors of \cite{Jang2017} did neither find conclusive answers on how to set $\lambda$, and used a constant value in some experiments, and an exponential (decaying) annealing scheme in others. 

Summarizing the most popular settings for $\lambda$ in practice, it can be seen that simply fixing its value between 0.1 and 1.0 is often done both for the GS estimator \cite{Mordatch2017,Grathwohl2018a,Wan2020,
Guo2020} and the ST-GS estimator \cite{Lu2017,Shetty,Xu2017,Choi2018a,Liu2018b,Bejnordi2019,Tjandra2019,Chonwiharnphan2020,Herrmann2020,Kang2020,Kojima2021}. Other works have annealed the temperature with various schemes from a value between 1.0 and 5.0 to a value in the range 0.1-0.5 \cite{Tsai2018,Grathwohl2018a,Huijben2019,Yang2019a,Yang2020a,Baevski2020, Wan2020, Guo2020}.

Finally, $\lambda$ could also be made trainable (or parameterized by a neural network), providing the model the freedom to learn the optimal entropy of the GS distribution (and the corresponding samples) \cite{Jang2017}. The authors of \cite{Havrylov, Yan2018} indeed both predict the (inverse) temperature with a (shallow) hyper-model. Both works propose a formula that yields a temperature between 0 and 1. 

\subsection{Interaction between \texorpdfstring{$\lambda$}{lambda} and \texorpdfstring{$T$}{T}}

\label{sec:T_and_lambda}

Section \ref{sec:boltzmann} already explained how the Boltzmann temperature $T$ yields a diversity-quality or exploration-exploitation trade-off, when sampling from a Boltzmann distribution (i.e. a tempered categorical). When the GS distribution is used to relax this categorical to enable gradient estimation, the two different temperatures (i.e. Boltzmann temperature $T$, and GS temperature $\lambda$) come into play. Figure \ref{fig:flat_draw_one_sample} illustrates the relations between relaxed and discrete samples from a categorical distribution, and shows how both temperatures relate. 

Equation (\ref{eq:scaled_gumbel_max}) in \cref{sec:gumbel_max} showed that sampling from a tempered categorical can also be achieved by scaling the Gumbel noise in the Gumbel-max trick. From track $B \veryshortarrow E$ in \cref{fig:flat_draw_one_sample} we now see that relaxing this approach with the GS distribution, e.g. done by \cite{Guo2020a}, causes an interaction effect between $\lambda$ and $T$, yielding a GS distribution with temperature $\lambda/T$.  

Instead of tuning the Boltzmann temperature to deal with the diversity-quality dilemma in the context of discrete GANs, the authors of \cite{Nie2019} leverage the GS temperature $\lambda$ instead. Reasoning is as follows; a high GS temperature $\lambda$ (low inverse temperature in their work), introduces a high bias in the GS gradient estimator. In order to (partly) mitigate this bias, the GAN's generator will implicitly be pushed towards predicting a sharp distribution (i.e. a distribution with a low implicit Boltzmann temperature $T$). Following this reasoning, a high $\lambda$ would thus encourage low $T$, i.e. exploitation, while a low $\lambda$ encourages exploration (large $T$), thanks to the implicit regulation of the generator.

\subsection{Sampled vs noise-free inference}

After having trained a discrete model with the (ST-)GS estimator or variants thereof, inference can be run in two distinct ways. Samples could either be drawn from the trained categorical, or the class with the highest probability can directly be selected (which is equal to the Gumbel-max trick on logits without Gumbel perturbations). The chosen strategy has not always been reported, but when reported mostly noise-free inference was used. Given that sampled inference does not require adjusted code for inference, we assume that when not reported, inference is typically done via sampling. Though, if sampling then takes place with the value for $\lambda$ at which training ended, this sampled inference still approaches noise-free inference in case $\lambda$ was annealed to a low value during training \cite{Wan2020}.

In the context of NAS, the inference strategy was much more often reported than in other applications, and it was always noise-free \cite{Louizos2018b,Chang2019,Herrmann2020,Kang2020,Cai2021}. Given the application, this seems a logical choice, as the final goal is typically to prune a neural network, while remaining with the best elements of the model  (indicated by the classes with the highest probabilities), rather than to sample a model architecture. Also in the context of (hard) attention, some authors have reported noise-free inference \cite{Abid2019,Niu2019}. The choice for noisy vs noise-free inference thus typically depends on the application. When for example the goal is to train a (discrete) generative model, it seems evident to sample from the trained distribution, rather than to use noise-free inference.

The authors of \cite{Herrmann2020} observe boosted inference results for ensembled predictions from sampling multiple times from the trained distribution over channel-gates. The same authors interestingly remark that during test time, the probabilistic gates could also be replaced by deterministic gates that are all 'on', in which the ST-GS estimator only serves as a (learned) regularizer during training. This view seems analogous to the way (non-learned) dropout is used as a regularization technique in neural network optimization. 

\section{Summary \& Future perspectives}
\label{sec:conclusion}
 
This review provides an overview of extensions and applications of the Gumbel-max trick in machine learning. The Gumbel-max trick \cite{Gumbel1954} is a method to draw an exact sample from an unnormalized categorical distribution. It was reviewed in \cref{sec:gumbel_max_trick}, and depicted in \cref{fig:flat_draw_one_sample} and \ref{fig:bottom_up_top_down}. The Gumbel-max trick closely links to well-known Poisson processes, and weighted reservoir sampling, as discussed in \cref{sec:uni_exp_gumbel_section}. Figure \ref{tab:uni_exp_gumbel} outlines the equivalences between these three concepts.

Currently, several extensions of the Gumbel-max trick have been proposed in the literature to create extended sampling algorithms, e.g. to sample from structured models. These algorithms were discussed in \cref{sec:extended_sampling_algorithms}, and visually related in \cref{tab:sampling}. The Gumbel-max trick gained much popularity thanks to a proposed relaxation based on the Gumbel-Softmax (GS) distribution \cite{Jang2017,Maddison2017} (see \cref{sec:gumbel_softmax}), that enables error backpropagation for training deep neural networks. The GS estimator and further developments in Gumbel-based gradient estimators were discussed in \cref{sec:gradient_estimators}. Applications of these estimators were discussed in \cref{sec:applications}, and categorized in \cref{fig:application_tree}. Section \ref{sec:considerations} discussed practical considerations and commonly-made choices regarding Gumbel-based algorithms in a machine learning context. 

Future directions can be found both from the application, as from the  theoretical perspective. At the application side, a large range of real-life problems could benefit from highly-expressive neural networks that incorporate a form of discrete stochasticity.
Drug research, for example, has for years been dominated by lab research, while investigations to drug-drug interactions \cite{Dai2020} and drug discovery \cite{Wieder2020} are currently moving forward with state-of-the-art machine learning models. The field of logistics is another big area that often deals with optimization over discrete spaces, making it suitable to exploit state-of-the-art discrete stochastic models. Moreover, the every-growing amount of acquired data poses challenges regarding their storage, transfer and processing. Modern radar systems acquire up to several terabytes of data per second, therewith saturating bandwidths and decelerating processing. Compression algorithms that achieve (extremely) high compression rates are therefore desirable. Data-driven neural compression models that perform (discrete) data selection and/or compression (including quantization) have already been found useful (see \cref{sec:applications}), but great advances in this field are still foreseen. 

From the theoretical perspective, we may expect (even more) efficient sampling algorithms to draw one or multiple samples from (unnormalized) structured models. We also observe a trend towards machine learning pipelines that increasingly leverage more structured components. These include layers that directly encode optimization problems \cite{Amos2017, Gould2021, Djolonga2017, Tschiatschek2018} or structured latent variables \cite{Grover2019, Xie2019, Mena2018, Paulus2020} that are tailored to a specific task at hand. We believe that the combination of discrete solvers and differentiable models \cite{Niculae2017,Lorberbom2019,Berthet2020} is promising and that the development of relaxed (Gumbel-based) gradient estimators will continue to be a fruitful area of research.

\ifCLASSOPTIONcompsoc
  \section*{Acknowledgments}
\else
  \section*{Acknowledgment}
\fi

I. Huijben gratefully acknowledges support from Onera Health and the project `OP-SLEEP'. The project `OP-SLEEP' is made possible by the European Regional Development Fund, in the context of OPZuid. W. Kool is funded by ORTEC. M. Paulus gratefully acknowledges support from the Max Planck ETH Center for Learning Systems. Resources used in preparing this research were provided, in part, by the Sustainable Chemical Processes through Catalysis (Suchcat) National Center of Competence in Research (NCCR).

\bibliography{references}
\bibliographystyle{IEEEtran}

\begin{IEEEbiography}[{\includegraphics[width=1\linewidth, clip]{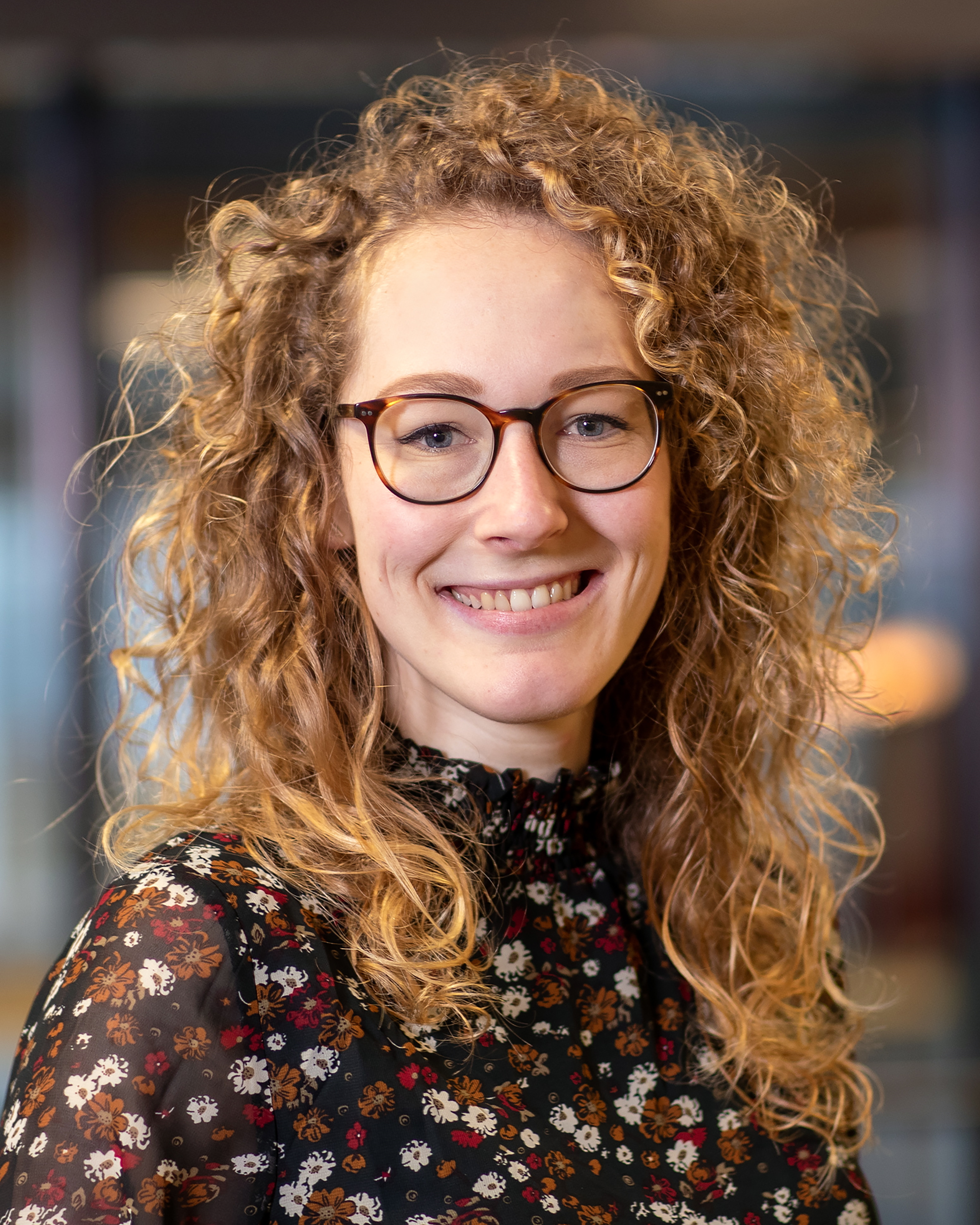}}]{Iris A. M. Huijben} (Student Member, IEEE) received her B.Sc. degree in Psychology \& Technology, and her M.Sc. degree in Electrical Engineering (both \emph{cum laude}), in which she enrolled as a PhD student afterwards. Both degrees were acquired from Eindhoven University of Technology, the Netherlands, where she is also pursuing her PhD currently. Her research interests include representation learning, compressed sensing and (biophysiological) signal processing. 
\end{IEEEbiography}

\begin{IEEEbiography}[{\includegraphics[width=1\linewidth,clip]{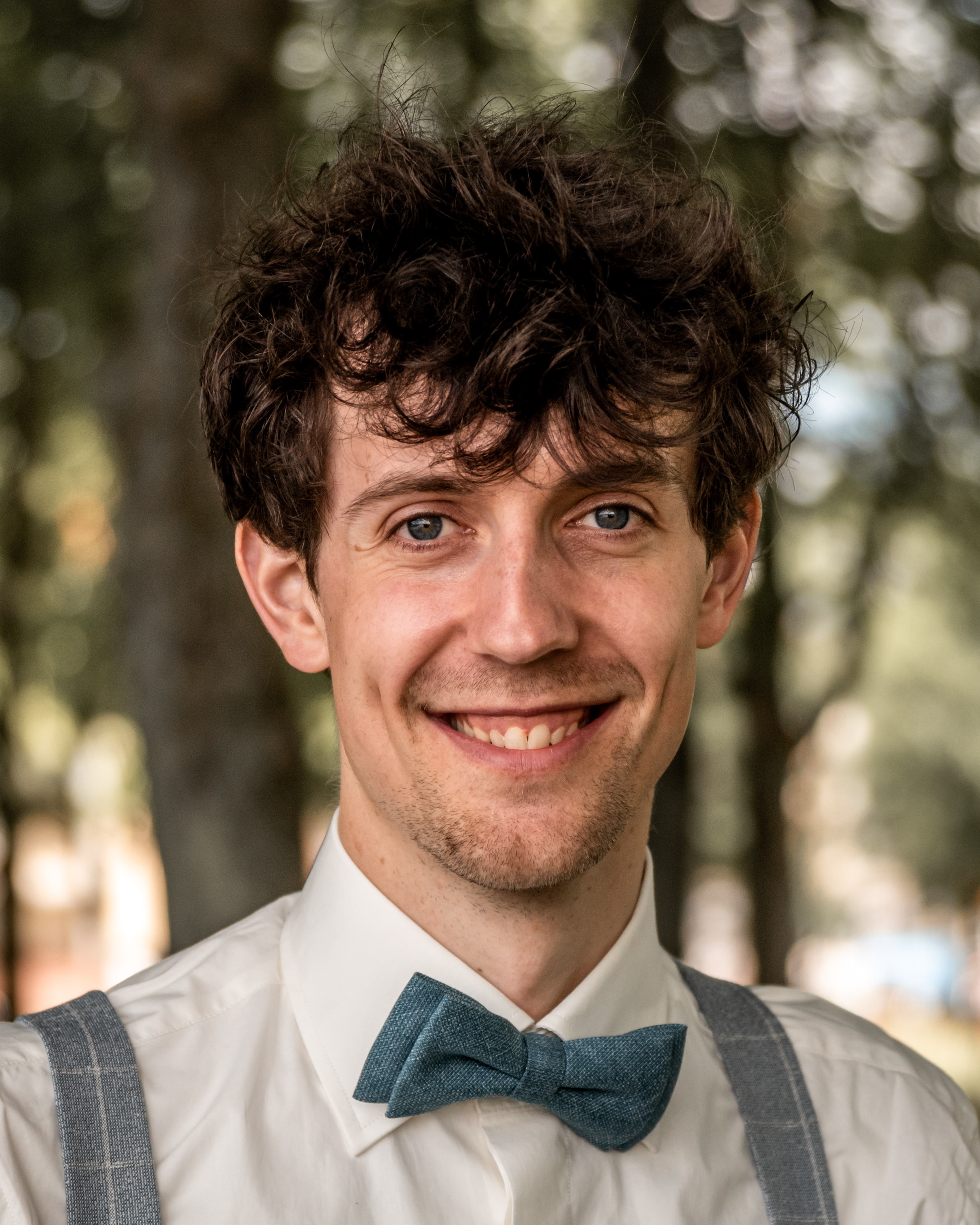}}]{Wouter Kool} received his M.Sc. degrees in Business Analytics and Operations Research (both \emph{cum laude}) from VU University in Amsterdam, The Netherlands. He is currently an Operations Research Engineer at ORTEC and a PhD student at the University of Amsterdam. His research focuses on theoretical and practical aspects of machine learning applied to combinatorial optimization.
\end{IEEEbiography}

\begin{IEEEbiography}[{\includegraphics[width=1\linewidth,clip]{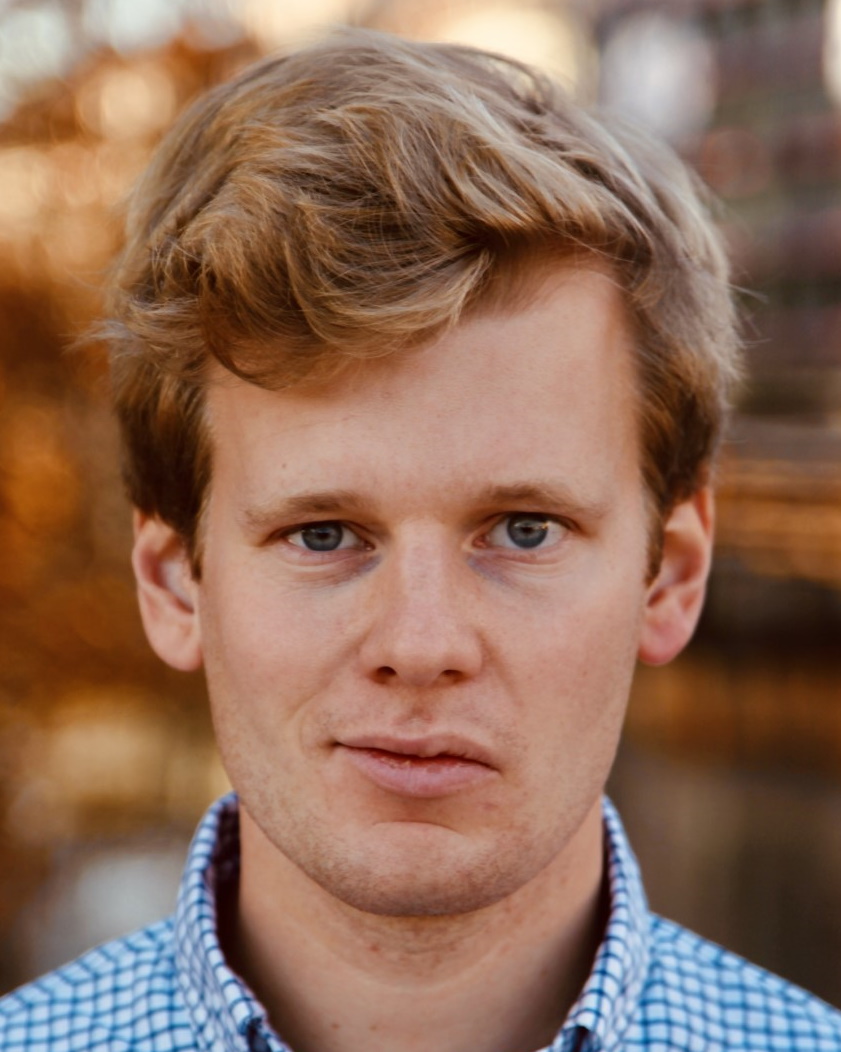}}]{Max B. Paulus} is a PhD student at ETH Zurich, a doctoral fellow of the Max Planck ETH Center for Learning Systems and an ELLIS PhD Student. He holds a M.Sc. degree in statistics from ETH Zurich and a B.Sc. degree (Hons) in economics from the University of Cambridge. His research interests lie at the intersection of machine learning and optimization and include latent variable models, interpretability in machine learning, as well as machine learning for integer programming.  
\end{IEEEbiography}

\begin{IEEEbiography}[{\includegraphics[width=1\linewidth,clip]{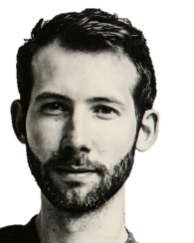}}]{Ruud J. G. van Sloun} (Member, IEEE) received the B.Sc. and M.Sc. degrees (\emph{cum laude}) in electrical engineering and the Ph.D. degree (\emph{cum laude}) from the Eindhoven University of Technology, Eindhoven, The Netherlands, in 2012, 2014, and 2018, respectively. Since then, he has been an Assistant Professor with the Department of Electrical Engineering at the Eindhoven University of Technology and since January 2020 a Kickstart-AI fellow at Philips Research, Eindhoven. From 2019-2020 he was also a Visiting Professor with the Department of Mathematics and Computer Science at the Weizmann Institute of Science, Rehovot, Israel. He is an NWO Rubicon laureate and received a Google Faculty Research Award in 2020. His current research interests include artificial intelligence and deep learning for front-end (ultrasound) signal processing, model-based deep learning, compressed sensing, ultrasound imaging, and probabilistic signal and image analysis.
\end{IEEEbiography}
\vfill

\newpage
\appendices
\section{Proof Gumbel-max trick and max-stability}
\label{app:gumbel_max_proof}

The Gumbel-max trick is used to draw an exact sample $\omega$ from a categorical distribution $\operatorname{Cat}(\boldsymbol{\theta})$, defined by
its unnormalized probabilities $\boldsymbol{\theta} \in \mathbb{R}_{\geq 0}^N$. The Gumbel-max trick \cite{Gumbel1954} takes the following form:

\begin{equation}
    \label{eq:gumbel_max_appendix}
    I = \underset{i \in D}{\operatorname{argmax}} \{\log \theta_i + G^{(i)} \} \sim{}  \operatorname{Cat}(\boldsymbol{\pi}),
\end{equation}
with $G^{(i)} \sim \operatorname{G}(0,1)$ i.i.d. $\forall i \in D$. We will now proof that $P[I=\omega] = \pi_{\omega}$, i.e. sampling using the Gumbel-max trick results in exact samples from $\operatorname{Cat}(\boldsymbol{\pi})$, with $\boldsymbol{\pi} = \boldsymbol{\theta} / \sum_{i \in D} \theta_i.$ We use $f_{\omega}(\cdot)$ to denote the PDF of $\operatorname{Gumbel}(\log \theta_{\omega})$, and introduce sub-domain $D' = D\backslash\{\omega\}$.

We start the proof with intuitively understanding that $I=\omega$ can only hold iff $G_{\log \theta_{\omega}} > G_{\log \theta_i} ~\forall i \in D'$. By independence of the (shifted) Gumbel variables, we can factorize the probability that all $G_{\log \theta_i}~(\text{with}~ i \in D')$, are smaller than $M := G_{\log \theta_{\omega}}$: 

\begin{align}
    \label{eq:gumbel_max_proof}
    &P[I=\omega] = \mathbb{E}_{M} \bigg[\prod_{i \in D'}p( G_{\log \theta_i} < M) \bigg] \nonumber \\
    &=\int_{-\infty}^{\infty} f_{\omega}(m)\prod_{i \in D'}p( G_{\log \theta_i} < m) ~dm  \nonumber \\
    &=\int_{-\infty}^{\infty}f_{\omega}(m)\prod_{i \in D'} e^{-e^{\log \theta_i - m}} ~dm \nonumber \\
    &=\int_{-\infty}^{\infty}f_{\omega}(m)~ e^{-\sum_{i \in D'}e^{\log \theta_i - m}} ~dm \nonumber \\
    &=\int_{-\infty}^{\infty} e^{\log \theta_{\omega}-m-e^{\log \theta_{\omega}-m}} e^{-\sum_{i \in D'}e^{\log \theta_i - m}} ~dm \nonumber \\
    &=\int_{-\infty}^{\infty} e^{\log \theta_{\omega}-m} e^{-\sum_{i \in D}e^{\log \theta_i - m}} ~dm  \nonumber\\
    &=\int_{-\infty}^{\infty} \theta_{\omega} e^{-m} e^{-e^{- m} \sum_{i \in D}\theta_i} ~dm.  \nonumber\\
    &~\nonumber \\
    & \mathrm{From~\cref{eq:normprobs_vs_unnorm_logits}:}~Z = \sum_{i \in D}\theta_i \mathrm{~and~} \theta_{\omega} = \pi_{\omega}Z: \nonumber \\
    &P[I=\omega] = \pi_{\omega}Z \int_{-\infty}^{\infty} e^{-m} e^{-Z e^{-m}} ~dm.
    &~\nonumber \\
    & \mathrm{Using} ~\int_{-\infty}^{\infty} e^{-m} e^{-Z e^{-m}} ~dm = \frac{1}{Z}:\nonumber  \\
    &P[I=\omega] = \pi_{\omega}. 
\end{align}

\noindent The max-stability property yields:

\begin{equation}
    M = \underset{i\in D}{\operatorname{max}}\{{\log \theta_i + G^{(i)}}\} \sim{} \operatorname{Gumbel}(\log Z).
\end{equation}

\noindent To proof that the maximum of a set of shifted Gumbels follows a Gumbel distribution with location $\log Z$, we formulate the joint distribution of the (independent) argmax and maximum. Since the (mathematical) steps are very similar to the proof from \cref{eq:gumbel_max_proof}, we deliberately omit some of them to prevent unnecessary clutter. Formally:

\begin{align}
    \label{eq:max_stability_proof}
    &p(M,I) =  f_{\omega}(m)\prod_{i \in D'}p(G_{\log \theta_i} < m) \nonumber \\
    &=\theta_{\omega}~e^{-m} e^{-e^{- m} \sum_{i \in D}\theta_i}  \nonumber\\
    &= \pi_{\omega} Z~e^{-m}e^{-Z e^{-m}} \nonumber\\
    &= \pi_{\omega}~f_{\log Z}(m)  \nonumber\\
    &= P[I=\omega]~p(M),
\end{align}
where \mbox{$p(M) = f_{\log Z}(m)$} denotes the PDF of a Gumbel variable located at $\log Z$, which thus implies that \mbox{$M \sim \operatorname{Gumbel}(\log Z)$}.

\section{Gumbel-max trick with non-standard Gumbel samples}
\label{app:gumbel_scaled_max_proof}

Given a set of i.i.d. standard Gumbel random variables $G^{(i)}$, their scaled and shifted counterparts $\beta G^{(i)} + \mu \sim \operatorname{G}(\mu,\beta)$ i.i.d. $\forall i \in D$, and the translation and scale \emph{invariance} of the argmax function, we can write:

\begin{alignat}{2}
    I' &=  \underset{i \in D}{\operatorname{argmax}} \{\log \theta_{i;T}  + \beta G^{(i)} + \mu \}  \nonumber\\
    &=  \underset{i \in D}{\operatorname{argmax}} \{a_i/T  + \beta G^{(i)} + \mu \}  \nonumber\\
    & = \underset{i \in D}{\operatorname{argmax}} \{a_i/T  + \beta G^{(i)}\} &{}\backslash \backslash{}&  \textit{translation~invariance}  \nonumber\\
    &= \underset{i \in D}{\operatorname{argmax}} \{a_i/(T\beta)  + G^{(i)}\}. &{}\backslash \backslash{}& \textit{scale~invariance} \nonumber
\end{alignat}

\noindent Given the Gumbel-max trick in \cref{eq:gumbel_max_appendix}, we conclude that $I' \sim \operatorname{Cat}(\frac{\exp(\va/(T\beta))}{\sum_{i \in D} \exp(a_i/(T\beta))}) = \operatorname{Cat}(\va, T\beta)$. \newline

\noindent Analogously, using the translation and scale \emph{equivariance} of the max function, we can write:
    
\begin{alignat}{2}
    M' &= \underset{i \in D}{\operatorname{max}} \{\log \theta_i  + \beta G^{(i)} + \mu\}  \nonumber \\
    &= \underset{i \in D}{\operatorname{max}} \{a_i/T  + \beta G^{(i)} + \mu\}  \nonumber \\
    &= \underset{i \in D}{\operatorname{max}} \{a_i/T  + \beta G^{(i)} \} + \mu &{}\backslash \backslash{}&  \textit{translation~equiv.} \nonumber \\
    &= \beta \cdot \underset{i \in D}{\operatorname{max}} \{a_i/(T\beta)  + G^{(i)}\} + \mu. &\hspace{0.3cm}{}\backslash \backslash{}&  \textit{scale~equiv.} \nonumber
\end{alignat}
Given the max-stability, scaling and shifting properties of Gumbels, we conclude that $M' \sim \operatorname{Gumbel}(\mu+\beta \log Z',\beta)$, with $Z' = \sum_{i \in D} \exp(a_i/(T\beta))$.

\section{Sampling experiments}
\label{app:sampling_experiments}

\begin{figure*}
    \centering
    \includegraphics[width=\linewidth,trim={2.7cm 0cm 6.5cm 0cm},clip]{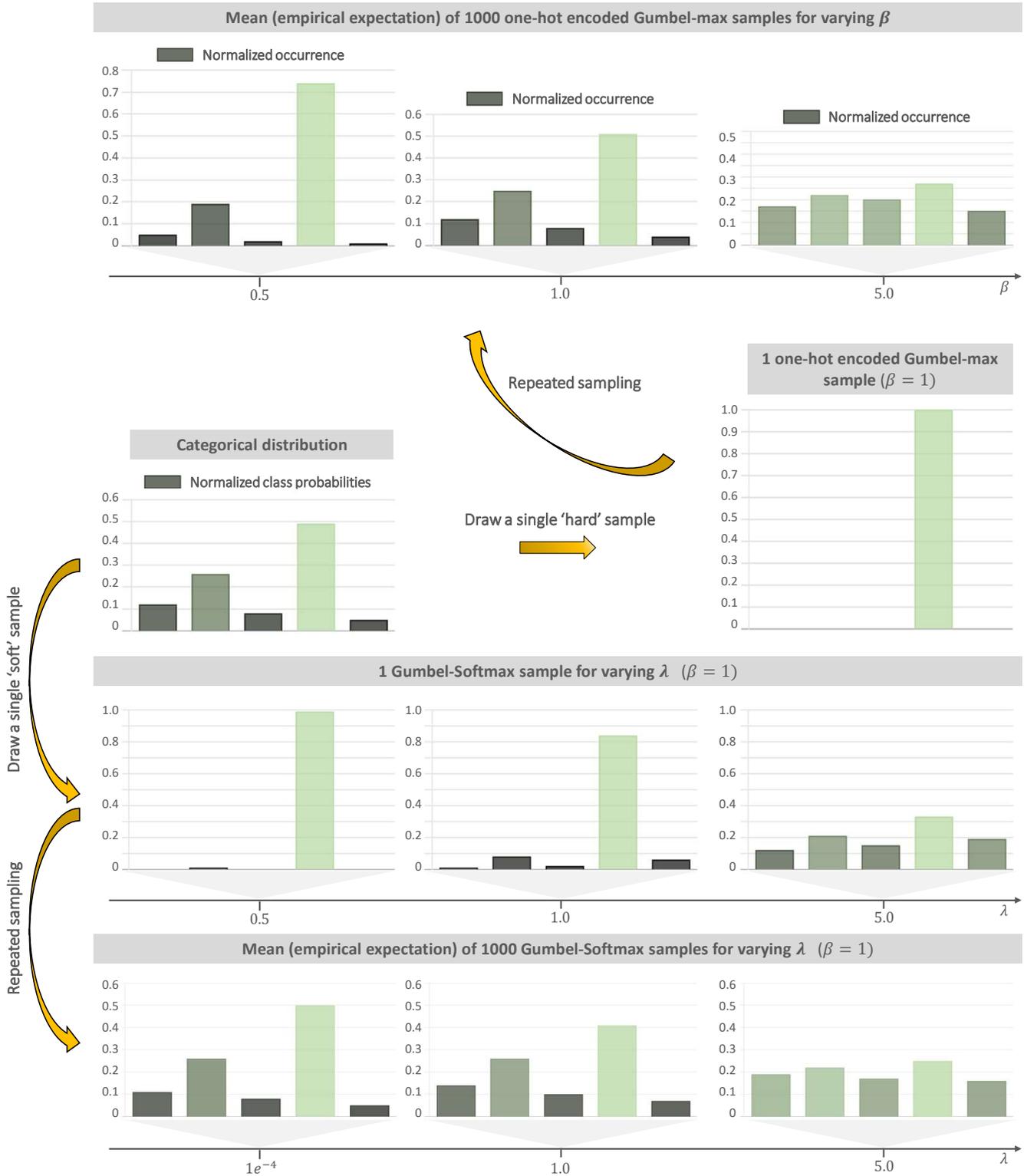}
    \caption{A categorical distribution with five classes was randomly created (2nd row, left), from which 1 Gumbel-max sample results in a one-hot vector (2nd row, right). When multiple (hard) samples are drawn, the (Monte-Carlo) histogram of the one-hot encoded Gumbel-max samples takes the form of the categorical distribution (1st row, middle). On the other hand, if the Gumbel noise's scale $\beta$ is altered from its default setting of 1, the distribution from which we effectively sample has lower entropy for $\beta < 1$ (1st row, left), or higher entropy (i.e. more uniform) for $\beta > 1$ (1st row, right). One GS sample `spreads' the mass over the different classes (as compared to the Gumbel-max sample), and varies for GS temperature $\lambda$ (3rd row). The expectation of multiple GS samples converges to the categorical distribution for $\lambda \rightarrow 0^+$, while it becomes more uniform for higher values of $\lambda$ (4rd row).}
    \label{fig:sampling_experiments}
\end{figure*}

Figure \ref{fig:sampling_experiments} shows the results of sampling experiments with varying values for Gumbel noise scale $\beta$ and GS temperature $\lambda$. Experiments were run with the demonstration setup that is provided with this paper, and can be found at \url{https://iamhuijben.github.io/gumbel_softmax_sampling.html}. 

\newpage
\section{Top-down sampling}
\label{app:top_down}
Algorithm \ref{alg:top_down} provides the top-down construction algorithm from \cite{Maddison2014a}, rewritten in discrete sampling space and with notations as used in the rest of this paper.

\begin{algorithm}[h!]
\caption{Top-Down construction}\label{alg:top_down}
\begin{algorithmic}
\Require Sampling space $D \in \{1,\ldots,N\}$, unnormalized probabilities $\boldsymbol{\theta} \in \mathbb{R}_{\geq 0}^N$, partition function $Z_D(\boldsymbol{\theta})= \sum_{i \in D} \theta_i$, queue $Q$
\State ~
\State $D_1 \leftarrow D$
\State $m_1 \sim \operatorname{Gumbel}(\log Z_{D_1}(\boldsymbol{\theta}))$
\State $\omega_1 \sim \operatorname{Cat}(\boldsymbol{\theta}/Z_{D_1}(\boldsymbol{\theta}))$
\State $Q.push(1)$
\State $k \leftarrow 1$
\While{$!Q.empty()$}
    \State $j \leftarrow Q.pop()$
    \State $L,R \leftarrow partition(D_j - \{\omega_j$\})
    \For{$C \in \{L,R\}$}
        \If{$C \neq \emptyset $}
            \State $k \leftarrow k+1$
            \State $D_k \leftarrow C$
            \State $m_k \sim \operatorname{TruncGumbel}(\log Z_{D_k}(\boldsymbol{\theta}),1,m_j)$
            \State $\omega_k \sim  \operatorname{Cat}(\frac{\mathbbm{1}(i \in D_k)\boldsymbol{\theta}}{Z_{D_k}(\boldsymbol{\theta})})$
            \State $Q.push(k)$ 
            \State \textbf{yield} ($m_k, \omega_k$)
        \EndIf
    \EndFor 
\EndWhile
\end{algorithmic}
\end{algorithm}

\end{document}